\documentclass[11pt]{article}

\usepackage[utf8]{inputenc}
\usepackage[T1]{fontenc}
\usepackage[margin=1in]{geometry}
\usepackage[round]{natbib}
\usepackage{hyperref}
\usepackage{url}
\usepackage{booktabs}
\usepackage{amsfonts}
\usepackage{amsmath}
\usepackage{amsthm}
\usepackage{nicefrac}
\usepackage{microtype}
\usepackage{xcolor}
\usepackage{graphicx}
\usepackage{algorithm}
\usepackage{algorithmic}
\usepackage{multirow}
\usepackage[normalem]{ulem}
\usepackage{subcaption}
\usepackage{placeins}
\usepackage{listings}
\usepackage{tikz}
\usepackage{pgfplots}
\pgfplotsset{compat=1.18}
\usetikzlibrary{arrows.meta,positioning,shapes.geometric,fit,backgrounds,
                 decorations.pathreplacing,calc,patterns}

\newtheorem{definition}{Definition}

\newtheorem{remark}{Remark}

\newcommand{\sg}{\textsc{HCG-RAG}}
\newcommand{\fgr}{\textsc{Fast-GraphRAG}}
\newcommand{\lgr}{\textsc{LightRAG}}
\newcommand{\nrag}{\textsc{NaiveRAG}}
\newcommand{\msgr}{\textsc{MS-GraphRAG}}
\newcommand{\epilepsyqa}{\textsc{EpilepsyQA}}
\newcommand{\spec}{\mathcal{S}}
\newcommand{\Dim}{\mathcal{A}}
\newcommand{\Metrics}{\mathcal{M}}
\newcommand{\Vars}{\mathcal{V}}
\newcommand{\Edges}{\mathcal{E}}
\newcommand{\Rels}{\mathcal{R}}
\newcommand{\Chunks}{\mathcal{C}}
\DeclareMathOperator{\cosim}{cos}

\title{Structure Over Scale: Schema-Constrained Causal Graphs for RAG}

\author{%
  Marc Saouda \quad Rajprakash Bale \quad Eren Aldis \quad Cloves Almeida \\
  Boston Consulting Group \\
  \texttt{saouda.marc@bcg.com \quad bale.rajprakash@bcg.com} \\
  \texttt{aldis.eren@bcg.com \quad almeida.cloves@bcg.com}
}

\date{}

\begin{document}

\maketitle

\begin{abstract}
Graph-based retrieval-augmented generation (GraphRAG) grounds answers
in structured knowledge, but current systems extract entities and
relationships exhaustively, producing graphs whose size and
construction cost scale with corpus length rather than with the
reasoning a query requires. We introduce \sg{} (Hierarchical Causal
Graph RAG), which replaces open-ended extraction with
\emph{schema-constrained causal graphs}: an automated pipeline
distills a corpus into a fixed, typed vocabulary of causal variables
and materializes a compact two-tier graph over it. Our
schema-constrained graphs match entity-relation baselines on answer
quality at a fraction of the cost: 3--20$\times$ fewer nodes,
8$\times$--135$\times$ fewer build-time LLM calls than the most
LLM-intensive baseline (\msgr{}), and graphs compact enough for a
domain expert to audit, correct, and extend. On medical and clinical
benchmarks, including a neurologist-validated epilepsy dataset, \sg{}
matches or exceeds the best entity-relation systems. An ablation
isolates the causal graph as a \emph{structured retrieval filter},
contributing $+$6 percentage points (pp) over embedding-only
retrieval. Across all domains with discoverable hierarchical causal
structure, only methods imposing higher-level organization outperform
flat entity-relation retrieval, indicating that \emph{what} is placed
in the graph matters more than how many nodes it contains.
\end{abstract}

\section{Introduction}
\label{sec:intro}

Retrieval-augmented generation (RAG) grounds LLM responses in external
knowledge, improving factuality and adaptability by separating
knowledge access from parametric memory~\citep{lewis2020rag,
guu2020realm, karpukhin2020dense, gao2024rag}. Modern systems retrieve
passages, fuse evidence across documents, decide when to retrieve, or
interleave retrieval with reasoning steps~\citep{izacard2021leveraging,
jiang2023active, trivedi2023interleaving, asai2024selfrag}. This is powerful but
largely unstructured: a top-$k$ retriever can surface relevant passages
without representing how concepts, mechanisms, and outcomes depend on
one another. The Graph RAG paradigm~\citep{edge2024local,
guo2024lightrag} addresses this gap by extracting entity-relation
graphs from the corpus for structured, multi-hop retrieval. In
practice, however, these systems extract entities exhaustively,
producing graphs whose size scales with corpus length rather than with
task-relevant structure: \msgr{}~\citep{edge2024local}, \lgr{}~\citep{guo2024lightrag}, and \fgr{}~\citep{circlemind2024fast} each build graphs
of 4{,}800--5{,}800 nodes on GraphRAG-Bench medical
(Table~\ref{tab:main-results}).

The scale problem is especially visible in specialized scientific and
clinical domains, where useful answers often require following
mechanistic links rather than merely retrieving co-mentioned entities.
Medical QA benchmarks and clinical LLM evaluations emphasize that
correct answers often depend on interventions, patient state, and
outcomes, not just topical overlap~\citep{pal2022medmcqa,
singhal2023large}. In such settings, a graph whose
nodes are every extracted entity can be both large and weakly aligned
with the reasoning variables that matter for retrieval. A smaller graph
can be preferable if its vocabulary is
restricted to the causal quantities a domain expert would recognize.

We ask whether this scale is necessary for strong retrieval. Building
on work showing that LLMs can efficiently discover causal
structure~\citep{jiralerspong2024efficient} and that causal graphs
improve RAG retrieval~\citep{wang2025causalrag}, we present \sg{}, a
system that constructs \emph{schema-constrained hierarchical causal
graphs} from unstructured corpora: an automated LLM-driven pipeline
produces a fixed, typed vocabulary of causal variables
(Definition~\ref{def:spec}), and two-tier graph construction
materializes graphs 3--20$\times$ smaller than entity-relation
baselines over the same corpus at 8$\times$--135$\times$ fewer
build-time LLM calls than \msgr{} (Appendix~\ref{app:cost}).
\emph{What} is put in the graph and how retrieval uses it
shift the quality--efficiency frontier.

We evaluate on three corpora: GraphRAG-Bench medical (2{,}062~Q),
GraphRAG-Bench novel (2{,}010~Q), and \epilepsyqa{}, a 375-question
benchmark that we curate from two open-access clinical epilepsy
textbooks and validate with two board-certified neurologists.
This design deliberately tests both the intended regime and a boundary
case: medical and epilepsy corpora contain hierarchical causal
structure, while literary narrative rewards broad entity and event
tracking. Schema-guided causal construction matches the best
entity-relation baselines on structured-knowledge corpora at a
fraction of the graph size, while entity-relation methods retain the
advantage on literary narrative, where discoverable hierarchical
causal structure is largely absent (Section~\ref{sec:results}).

We make three contributions:\footnote{Code,
evaluation scripts, \epilepsyqa{} questions and annotations,
and corpus reconstruction instructions are available at
\url{https://anonymous.4open.science/r/hcg-rag}.}
(i)~\textbf{compact, inspectable graphs at
competitive quality}: schema-constrained construction produces
graphs 3--20$\times$ smaller than entity-relation GraphRAG on medical
and epilepsy, small enough for a domain expert to audit and
correct while matching answer quality, via a three-stage
automated pipeline whose query path is one LLM call plus two embedding
calls;
(ii)~\textbf{structure over scale on specialized domains}: only
retrieval methods that impose higher-level organization outperform
flat entity-relation retrieval, and schema-guided causal graphs are
a lean, high-precision instance;
(iii)~\textbf{\epilepsyqa{}}, a curated clinical epilepsy benchmark
of 375~questions spanning six reasoning types, derived from two
open-access textbooks and validated by two board-certified
neurologists, released as a public resource for evaluating
domain-specialized retrieval systems
(Sections~\ref{sec:results},~\ref{sec:ablation}).

\section{Related work}
\label{sec:related}

\paragraph{Retrieval-augmented generation.}
Open-domain QA and retrieval-augmented language modeling established
the basic pattern of retrieving external evidence before
generation~\citep{karpukhin2020dense, guu2020realm, lewis2020rag}.
Subsequent work improved how retrieved text is consumed, from
fusion-in-decoder architectures~\citep{izacard2021leveraging} to
systems that decide when to retrieve, critique retrieved evidence, or
interleave retrieval with multi-step reasoning~\citep{jiang2023active,
trivedi2023interleaving, asai2024selfrag}. These approaches treat the
retrieval corpus primarily as text. \sg{} instead builds a typed graph
offline and uses it only to select evidence passages, leaving the
answer LLM with textual support rather than graph triples.

\paragraph{Graph-based retrieval-augmented generation.}
GraphRAG systems vary along three axes: \emph{extraction strategy}
(unconstrained entity extraction vs.\ schema-guided construction),
\emph{graph organization} (flat entity-relation graphs vs.\
hierarchical structures such as community summaries or typed schemas),
and \emph{retrieval mechanism} (key-value lookup, personalized
PageRank, community-level map-reduce, or graph traversal). We compare
\sg{} to three representative baselines that span these axes.
\msgr{}~\citep{edge2024local} extracts entity-relation triples and
applies Leiden community detection with hierarchical summaries for
local and global search; \lgr{}~\citep{guo2024lightrag} adds
dual-level key-value retrieval over an LLM-extracted entity-relation
graph; and \fgr{}~\citep{circlemind2024fast} replaces community
detection with personalized PageRank. All three share
\emph{unconstrained entity extraction}, producing large heterogeneous
graphs whose structure mirrors surface-level mentions. Earlier
graph-and-text QA systems also showed the value of traversing
structured connections alongside passages~\citep{sun2019pullnet,
oguz2022unikqa}, but typically assume a knowledge graph or learn a
unified retriever over existing graph/text sources. \sg{} replaces
unconstrained entity extraction with schema-guided causal construction,
yielding graphs 3--20$\times$ smaller that nonetheless capture the
causal structure needed for complex reasoning. Relative to the current
GraphRAG-Bench leaders, our target is therefore a different operating
point: fewer, typed, auditable graph nodes at comparable quality rather
than the highest score under an unconstrained graph budget.

\paragraph{Knowledge graphs and LLMs.}
Knowledge-graph augmentation can improve LLM factuality and reasoning
by injecting triples, subgraphs, or graph-derived textual context into
prompts~\citep{baek2023kaping, pan2024kgllm}. Recent systems extend
this idea to textual graph understanding, for example by retrieving
graph context for generation~\citep{he2024gretriever}. These methods
primarily ask how to use an
available structured resource at inference time. Our setting differs in
two ways: the graph is constructed from unstructured corpora, and its
schema is constrained before graph construction so that nodes denote
typed variables rather than arbitrary entities.

\paragraph{LLM-based causal discovery and causal RAG.}
Classical causal discovery
infers graph structure from data under statistical assumptions, while
recent surveys emphasize both the promise and fragility of such
methods in complex domains~\citep{glymour2019review}. LLMs add a
different source of signal: they encode background knowledge that can
support causal reasoning, but benchmarks also show that this ability is
uneven and must be constrained~\citep{kiciman2024causal,
jin2023cladder}. \citet{jiralerspong2024efficient} discover a causal
directed acyclic graph (DAG) over a \emph{predefined} variable set via
breadth-first search (BFS)--based querying, and
\citet{wang2025causalrag} add per-query LLM calls to trace causal paths
within an entity-relation graph built by standard extraction. \sg{}
pushes causal reasoning to \emph{construction time}: a typed causal
schema (Definition~\ref{def:spec}) discovers both variables and
relationships, the fixed metric vocabulary constrains the graph to
causally relevant concepts by design, and query-time retrieval requires
only embedding matching and BFS traversal (no per-query LLM causal
reasoning). The novelty is this combination of automated variable
discovery, schema-constrained causal graph construction, and
evidence-only retrieval; prior causal-RAG systems either assume a
predefined variable set or reason over causal paths at query time.

\section{Method}
\label{sec:method}

\subsection{Problem setup}
\label{sec:problem-setup}

We study retrieval-augmented question answering over an unstructured
corpus $\mathcal{D} = \{d_i\}_{i=1}^n$ in domains where answers often
depend on causal mechanisms. The system is given an \emph{outcome
configuration} $\mathcal{O}$ specifying graph roles, including
terminal sink metrics (e.g., ``Patient Outcome'') and retrieval
outcome metrics that designate endpoint nodes for graph-guided evidence
selection. Terminal metrics receive
causal edges but emit none; nonterminal outcome metrics may still
participate in upstream or downstream mechanisms, but are treated as
retrieval endpoints when reached. For a query $q$, the method constructs a graph
$G = (\Vars, \Edges, \phi)$ with variable nodes $\Vars$, directed
causal edges $\Edges$, and an evidence map
$\phi:\Edges \to 2^{\Chunks}$ that links each edge to supporting
chunks from the chunk-level partition $\Chunks = \{c_j\}$ of
$\mathcal{D}$. A retrieval policy $R_G(q) \subseteq \Chunks$ selects
evidence
from $G$, and the answer model returns
$a = \mathrm{LLM}(q, R_G(q))$. We evaluate \sg{} by the joint
trade-off among answer correctness, graph compactness, and build/query
cost: smaller graphs are human-auditable and enable expert-in-the-loop
curation, while lower API cost widens the deployment envelope.

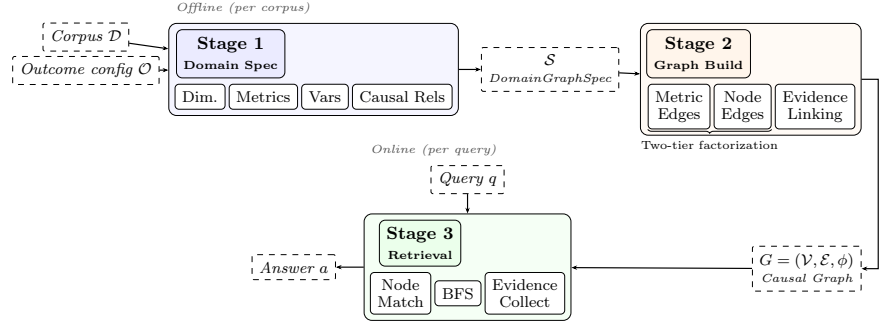
\begin{figure*}[!ht]
\centering
\resizebox{0.7\linewidth}{!}{
\begin{tikzpicture}[
    stage/.style={draw, rounded corners=3pt, minimum height=0.6cm,
                  minimum width=1.4cm, font=\footnotesize\bfseries, align=center},
    substep/.style={draw, rounded corners=2pt, minimum height=0.38cm,
                    font=\scriptsize, align=center, fill=white},
    artifact/.style={draw, rounded corners=1pt, dashed,
                     minimum height=0.38cm, font=\scriptsize\itshape,
                     align=center},
    arr/.style={-{Stealth[length=3pt]}, semithick},
    lbl/.style={font=\tiny, text=black!70},
]


\node[artifact] (corpus) at (0, 0.3) {Corpus $\mathcal{D}$};
\node[artifact] (config) at (0, -0.3) {Outcome config $\mathcal{O}$};

\node[stage, fill=blue!8] (s1) at (2.6, 0) {Stage 1\\[-1pt]\tiny Domain Spec};
\node[substep, below=0.1cm of s1, xshift=-0.55cm] (dim) {\scriptsize Dim.};
\node[substep, right=0.06cm of dim] (met) {\scriptsize Metrics};
\node[substep, right=0.06cm of met] (var) {\scriptsize Vars};
\node[substep, right=0.06cm of var] (rel) {\scriptsize Causal Rels};

\begin{scope}[on background layer]
  \node[draw, rounded corners=4pt, fill=blue!4, inner sep=3pt,
        fit=(s1)(dim)(met)(var)(rel)] (s1box) {};
\end{scope}

\draw[arr] (corpus) -- (s1box);
\draw[arr] (config) -- (s1box);

\node[artifact, right=0.4cm of s1box] (spec)
  {$\spec$\\[-1pt]\tiny DomainGraphSpec};
\draw[arr] (s1box) -- (spec);

\node[stage, fill=orange!8] (s2) at (11.0, 0) {Stage 2\\[-1pt]\tiny Graph Build};
\node[substep, below=0.1cm of s2, xshift=-0.35cm] (me) {\scriptsize Metric\\[-1pt]\scriptsize Edges};
\node[substep, right=0.06cm of me] (ne) {\scriptsize Node\\[-1pt]\scriptsize Edges};
\node[substep, right=0.06cm of ne] (ev) {\scriptsize Evidence\\[-1pt]\scriptsize Linking};

\begin{scope}[on background layer]
  \node[draw, rounded corners=4pt, fill=orange!4, inner sep=3pt,
        fit=(s2)(me)(ne)(ev)] (s2box) {};
\end{scope}

\draw[arr] (spec) -- (s2box);

\draw[decorate, decoration={brace, amplitude=2pt, mirror}]
  (me.south west) -- (ne.south east)
  node[midway, below=2pt, font=\tiny] {Two-tier factorization};

\node[lbl, above=0.03cm of s1box.north west, anchor=south west]
  {\textit{Offline (per corpus)}};

\draw[arr] (s2box.east) -- ++(0.3, 0) |- ++(0, -3.4) -- ++(-0.3, 0)
  coordinate (turnend);

\node[artifact, anchor=east] (graph) at (turnend)
  {$G = (\Vars, \Edges, \phi)$\\[-1pt]\tiny Causal Graph};


\node[stage, fill=green!8] (s3) at (6.0, -3.4) {Stage 3\\[-1pt]\tiny Retrieval};
\node[substep, below=0.1cm of s3, xshift=-0.35cm] (match) {\scriptsize Node\\[-1pt]\scriptsize Match};
\node[substep, right=0.06cm of match] (bfs) {\scriptsize BFS};
\node[substep, right=0.06cm of bfs] (coll) {\scriptsize Evidence\\[-1pt]\scriptsize Collect};

\begin{scope}[on background layer]
  \node[draw, rounded corners=4pt, fill=green!4, inner sep=3pt,
        fit=(s3)(match)(bfs)(coll)] (s3box) {};
\end{scope}

\draw[arr] (graph) -- (s3box);

\node[artifact, above=0.35cm of s3box] (query) {Query $q$};
\draw[arr] (query) -- (s3box);

\node[lbl, above=0.03cm of query.north east, anchor=south east]
  {\textit{Online (per query)}};

\node[artifact, left=0.5cm of s3box] (ans) {Answer $a$};
\draw[arr] (s3box) -- (ans);

\end{tikzpicture}}
\caption{The \sg{} pipeline. The method takes a corpus and an outcome
configuration specifying the causal sinks the graph should build
toward (e.g., ``Patient Outcome'').
\textbf{Stage~1} generates a causal domain specification $\spec$ via
LLM-driven generation of dimensions, metrics, variables, and causal
relations, anchored by the configured outcome hierarchy. \textbf{Stage~2}
builds the causal graph through two-tier edge discovery (metric-level
templates then node-level instantiation) with evidence linking.
\textbf{Stage~3} answers queries via embedding-based node matching,
bounded BFS traversal, and evidence collection. Stages~1--2 run offline
per corpus; Stage~3 runs online per query.}
\label{fig:pipeline}
\end{figure*}

\sg{} constructs hierarchical causal graphs from unstructured corpora
and uses them for graph-guided evidence retrieval at query time.
Given $(\mathcal{D}, \mathcal{O})$, it proceeds in three stages
(Figure~\ref{fig:pipeline}): automated domain-specification generation
(\S\ref{sec:domspec}), two-tier graph build
(\S\ref{sec:graph-build}), and graph-guided retrieval
(\S\ref{sec:retrieval}); Algorithm~\ref{alg:pipeline}
(Appendix~\ref{app:algorithm}) gives the full formal specification.

Informally, a specification organizes the domain along four concepts.
\emph{Dimensions} are categorical axes that stratify the domain
(e.g., disease type, anatomical site). \emph{Metrics} are measurable
or assessable quantities along those axes (e.g., 5-year survival
rate). \emph{Variables} pair a metric with specific dimension
entities to produce a concrete measurable quantity (e.g., ``efficacy
of chemotherapy for NSCLC''). \emph{Seed edges} are directed causal
relationships between variables, annotated with mechanism
descriptions.

\begin{definition}[Domain Graph Specification]
\label{def:spec}
A domain graph specification is a tuple
$\spec = (\Dim, \Metrics, \Vars_0, \Rels_0)$ where:
\begin{itemize}
\item $\Dim = \{(\delta_i, E_i)\}$ is a set of dimensions, each a named
  axis $\delta_i$ with entity set $E_i$;
\item $\Metrics$ is a \textbf{fixed} set of typed metrics, each with a
  dimension signature $\operatorname{sig}(m) \subseteq \{\delta_i\}$ specifying the
  dimensions it binds;
\item $\Vars_0$ is an \textbf{extensible} seed set of variables, where each
  $v = (m, \mathbf{e})$ pairs a metric $m \in \Metrics$ with a tuple of
  dimension entities $\mathbf{e} \in \prod_{\delta_i \in \operatorname{sig}(m)} E_i$,
  producing a concrete
  measurable quantity
  (e.g., $v = (m_{\text{efficacy}}, [\text{chemo}, \text{NSCLC}])$);
\item $\Rels_0 \subseteq \Vars_0 \times \Vars_0$ is an extensible set of
  directed causal seed edges.
\end{itemize}
Fixed sets are exhaustive: graph building must only reference defined
members. Extensible sets are exemplary: graph building expands them as
corpus evidence warrants, constrained by the fixed vocabulary.
\end{definition}

\FloatBarrier
\begin{figure}[!ht]
\centering
\resizebox{0.7\linewidth}{!}{%
\begin{tikzpicture}[
    closed/.style={draw, thick, rounded corners=3pt, fill=blue!6,
                   minimum height=0.7cm, font=\scriptsize, align=center,
                   inner sep=5pt},
    open/.style={draw, dashed, rounded corners=3pt, fill=orange!6,
                 minimum height=0.7cm, font=\scriptsize, align=center,
                 inner sep=5pt},
    arr/.style={-{Stealth[length=3pt]}, thick},
    lbl/.style={font=\tiny, text=black!60, inner sep=1.5pt,
                fill=white, fill opacity=0.85, text opacity=1},
    tag/.style={font=\tiny\bfseries, rounded corners=1pt, inner sep=1.8pt},
    closedtag/.style={tag, fill=blue!15, text=blue!70},
    opentag/.style={tag, fill=orange!15, text=orange!70},
]

\node[closed] (dim) at (0, 2.7)
  {Dimensions\\[-1pt]\tiny e.g.\ disease type, treatment};
\node[closedtag] at ([xshift=0.6cm]dim.north west) {FIXED};

\node[open] (ent) at (5.6, 2.7)
  {Entities\\[-1pt]\tiny e.g.\ NSCLC, chemotherapy};
\node[opentag] at ([xshift=0.5cm]ent.north west) {EXT.};

\draw[arr] (dim) -- node[above, lbl] {contain} (ent);

\node[closed] (met) at (0, 0.6)
  {Metrics\\[-1pt]\tiny e.g.\ survival rate, treatment efficacy};
\node[closedtag] at ([xshift=0.6cm]met.north west) {FIXED};

\node[open, minimum width=3.0cm] (var) at (5.6, 0.6)
  {Variables\\[-1pt]\tiny e.g.\ SurvivalRate(NSCLC),\\[-1pt]\tiny Efficacy(NSCLC, chemo)};
\node[opentag] at ([xshift=0.5cm]var.north west) {EXT.};

\draw[arr] (met.east) -- node[above, lbl] {typed by} (var.west);
\draw[arr] (ent.south) -- node[right=3pt, lbl] {bound to} (var.north);

\node[open, minimum width=3.0cm] (rel) at (5.6, -1.9)
  {Causal Relations\\[-1pt]\tiny e.g.\ Efficacy(NSCLC, chemo)\\[-1pt]\tiny $\to$ SurvivalRate(NSCLC)};
\node[opentag] at ([xshift=0.5cm]rel.north west) {EXT.};

\draw[arr] (rel.north) -- node[right=3pt, lbl] {between} (var.south);

\node[draw, rounded corners=3pt, fill=green!8, minimum height=0.7cm,
      minimum width=1.55cm, font=\scriptsize, align=center] (graph) at (9.7, -0.65)
  {Causal\\Graph $G$};

\draw[arr] (var.east) -- node[above, lbl, pos=0.55] {nodes} (graph.west);
\draw[arr] (rel.east) -- node[below, lbl, pos=0.55] {edges} (graph.west);

\node[closed, minimum height=0.3cm, minimum width=0.5cm] at (0.6, -3.0) {};
\node[lbl, anchor=west, fill=none] at (1.0, -3.0) {fixed (exhaustive vocabulary)};
\node[open, minimum height=0.3cm, minimum width=0.5cm] at (5.4, -3.0) {};
\node[lbl, anchor=west, fill=none] at (5.8, -3.0) {extensible (expandable by graph build)};

\end{tikzpicture}}
\caption{Structure of the domain graph specification $\spec$.
Dimensions and metrics form a \textbf{fixed} vocabulary (solid borders):
graph building must use only defined types. Entities, variables, and causal
relations are \textbf{extensible} (dashed borders): the specification
provides seed examples that graph building expands as corpus evidence
warrants.}
\label{fig:domspec}
\end{figure}
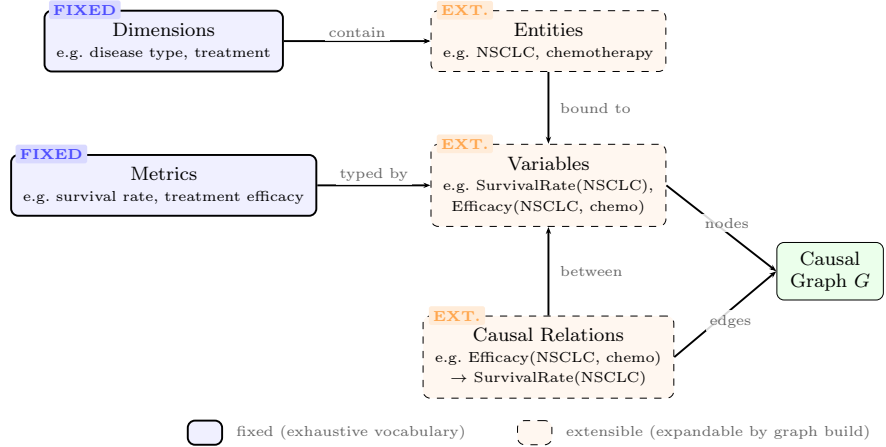

\subsection{Stage 1: Automated domain-specification generation}
\label{sec:domspec}

Given corpus $\mathcal{D}$, a chunk-wise LLM pipeline produces
$\spec$ (Definition~\ref{def:spec}; Figure~\ref{fig:domspec}) that
constrains all subsequent graph building. Each metric is typed as \textsc{Fixed} (exhaustive) or
\textsc{Extensible}; the LLM discovers dimensions, metrics,
variables, and directed causal seed edges from corpus chunks.
A deterministic assembly step drops invalid references, fuzzy-matches
relation endpoints to repair naming drift, and tags variables with
their configured roles.

\subsection{Stage 2: Two-tier causal graph build}
\label{sec:graph-build}

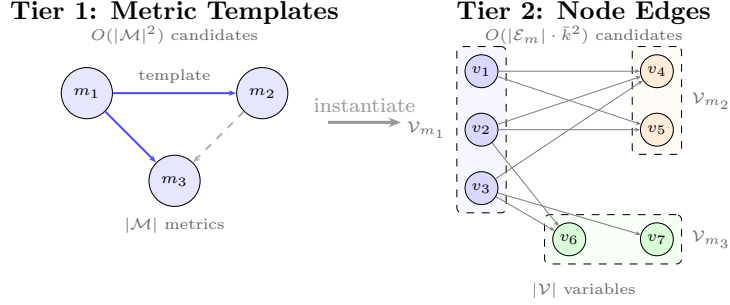
\begin{figure}[!ht]
\centering\resizebox{0.6\linewidth}{!}{
\begin{tikzpicture}[
    mnode/.style={draw, circle, minimum size=0.7cm, font=\tiny\bfseries,
                  fill=blue!10, inner sep=1pt},
    vnode/.style={draw, circle, minimum size=0.45cm, font=\tiny,
                  fill=orange!10, inner sep=0pt},
    arr/.style={-{Stealth[length=3pt]}, thick},
    darr/.style={-{Stealth[length=2pt]}, thin, black!50},
    lbl/.style={font=\tiny, text=black!60},
]

\node[font=\small\bfseries] at (1.2, 2.3) {Tier 1: Metric Templates};
\node[lbl] at (1.2, 2.0) {$O(|\Metrics|^2)$ candidates};

\node[mnode] (m1) at (0, 1.2) {$m_1$};
\node[mnode] (m2) at (2.4, 1.2) {$m_2$};
\node[mnode] (m3) at (1.2, 0) {$m_3$};

\draw[arr, blue!70] (m1) -- node[above, lbl] {template} (m2);
\draw[arr, blue!70] (m1) -- (m3);
\draw[arr, blue!70, dashed, black!30] (m2) -- (m3);

\node[lbl] at (1.2, -0.6) {$|\Metrics|$ metrics};

\draw[-{Stealth[length=5pt]}, very thick, black!40]
  (3.3, 0.8) -- node[above, font=\scriptsize] {instantiate} (4.3, 0.8);

\node[font=\small\bfseries] at (6.8, 2.3) {Tier 2: Node Edges};
\node[lbl] at (6.8, 2.0) {$O(|\Edges_m| \cdot \bar{k}^2)$ candidates};

\node[vnode, fill=blue!15] (v11) at (5.4, 1.5) {$v_1$};
\node[vnode, fill=blue!15] (v12) at (5.4, 0.7) {$v_2$};
\node[vnode, fill=blue!15] (v13) at (5.4, -0.1) {$v_3$};

\node[vnode, fill=orange!15] (v21) at (7.8, 1.5) {$v_4$};
\node[vnode, fill=orange!15] (v22) at (7.8, 0.7) {$v_5$};

\node[vnode, fill=green!15] (v31) at (6.6, -0.8) {$v_6$};
\node[vnode, fill=green!15] (v32) at (7.8, -0.8) {$v_7$};

\draw[darr] (v11) -- (v21);
\draw[darr] (v11) -- (v22);
\draw[darr] (v12) -- (v21);
\draw[darr] (v12) -- (v22);
\draw[darr] (v13) -- (v21);

\draw[darr] (v12) -- (v31);
\draw[darr] (v13) -- (v31);
\draw[darr] (v13) -- (v32);

\begin{scope}[on background layer]
  \node[draw, dashed, rounded corners=2pt, fill=blue!3,
        fit=(v11)(v12)(v13), inner sep=3pt, label={[lbl]left:$\Vars_{m_1}$}] {};
  \node[draw, dashed, rounded corners=2pt, fill=orange!3,
        fit=(v21)(v22), inner sep=3pt, label={[lbl]right:$\Vars_{m_2}$}] {};
  \node[draw, dashed, rounded corners=2pt, fill=green!3,
        fit=(v31)(v32), inner sep=3pt, label={[lbl]right:$\Vars_{m_3}$}] {};
\end{scope}

\node[lbl] at (6.8, -1.5) {$|\Vars|$ variables};

\end{tikzpicture}}
\caption{Two-tier edge construction. \textbf{Tier~1} discovers causal
templates between metric pairs (left). \textbf{Tier~2} instantiates
each template between the concrete variables belonging to the source
and target metrics (right). This factorization reduces LLM calls from
$O(|\Vars|^2)$ candidate pairs to
$O(|\Metrics|^2 + |\Edges_m| \cdot \bar{k}^2)$
(Remark~\ref{rem:search-space}).}
\label{fig:two-tier}
\end{figure}

Recall $G = (\Vars, \Edges, \phi)$ from
Section~\ref{sec:problem-setup}. Construction proceeds in two tiers
(Figure~\ref{fig:two-tier}). \textbf{Tier~1} evaluates each ordered
metric pair $(m_s, m_t) \in \Metrics \times \Metrics$ (terminals
excluded as sources) using evidence summaries and optional semantic
search, producing metric-level edge templates $\Edges_m$.
\textbf{Tier~2} instantiates each template between the concrete
variables of its source and target metrics; the LLM evaluates each
candidate edge, producing the final edge set $\Edges$ with mechanism
text and confidence scores. Each edge is linked to corpus chunks via
substring text matching plus semantic matching at threshold
$\tau_e{=}0.78$ (Appendix~\ref{app:evidence}).

\begin{remark}[Search-space reduction]
\label{rem:search-space}
Tier~1 prunes most metric pairs, so Tier~2 evaluates far fewer
candidate edges than exhaustive pairwise comparison. Formally, let
$\bar{k} = \max_{m \in \Metrics} |\Vars_m|$; the two-tier
construction requires at most $O(|\Metrics|^2 + |\Edges_m| \cdot
\bar{k}^2)$ LLM calls versus $\Theta(|\Vars|^2)$ for all-pairs
evaluation, with $|\Edges_m| \ll |\Metrics|^2$ in practice
(derivation in Appendix~\ref{app:algorithm};
Figure~\ref{fig:two-tier}).
\end{remark}

\begin{remark}[Auditable evidence]
\label{rem:auditable}
$\spec$ is a compact YAML (Figure~\ref{fig:domspec}); edits
propagate deterministically
through Stages~2--3. Each node $v = (m, \mathbf{e})$ carries a typed
metric and dimension-entity binding, each edge in $\Edges$ carries
its Stage-1 mechanism text, and $\phi$ links the edge back to
specific corpus chunks, so any retrieved evidence is traceable to its
source. All experiments use fully automated specifications.
\end{remark}

\subsection{Stage 3: Graph-guided evidence retrieval}
\label{sec:retrieval}

\begin{figure}[!ht]
\centering\resizebox{0.7\linewidth}{!}{
\begin{tikzpicture}[
    qnode/.style={draw, rounded corners=2pt, fill=yellow!20,
                  minimum height=0.5cm, font=\scriptsize, align=center},
    gnode/.style={draw, circle, minimum size=0.5cm, font=\tiny,
                  inner sep=1pt},
    matched/.style={gnode, fill=green!25, line width=1pt},
    reached/.style={gnode, fill=blue!10},
    outcome/.style={gnode, fill=red!15, double},
    unreach/.style={gnode, fill=black!5, draw=black!30, text=black!40},
    evid/.style={draw, rounded corners=1pt, fill=orange!8,
                 font=\tiny, align=left, text width=2.2cm},
    arr/.style={-{Stealth[length=2.5pt]}, thick},
    garr/.style={-{Stealth[length=2pt]}, thin, black!40},
    lbl/.style={font=\tiny, text=black!60, align=center},
]

\node[qnode] (q) at (0, 3.0) {Query: ``Reed-Sternberg\\cells, diagnosis?''};

\node[matched] (n1) at (0, 1.2) {$v_1$};
\node[lbl, left=0.1cm of n1] {diagnostic\\markers};
\node[reached] (n2) at (1.8, 1.8) {$v_2$};
\node[lbl, above=0.02cm of n2] {cell morph.};
\node[reached] (n3) at (1.8, 0.6) {$v_3$};
\node[lbl, below=0.02cm of n3] {lymphoma\\subtype};
\node[outcome] (n4) at (3.6, 1.2) {$v_4$};
\node[lbl, above=0.05cm of n4] {disease\\diagnosis};
\node[matched] (n5) at (0, 0) {$v_5$};
\node[lbl, left=0.1cm of n5] {histopath.\\findings};
\node[unreach] (n6) at (1.8, -0.6) {$v_6$};
\node[unreach] (n7) at (3.6, -0.6) {$v_7$};

\draw[arr, green!60!black, dashed] (q) -- node[right, lbl] {$\cosim \geq \tau$} (n1);

\draw[arr, blue!70] (n1) -- (n2);
\draw[arr, blue!70] (n1) -- (n3);
\draw[arr, blue!70] (n2) -- (n4);
\draw[arr, blue!70] (n3) -- (n4);
\draw[arr, blue!70] (n5) -- (n3);
\draw[garr] (n6) -- (n7);

\node[evid] (e1) at (6.2, 2.2) {\textit{``RS cells are pathog-}\\
\textit{nomonic for Hodgkin}\\
\textit{lymphoma\ldots''}};
\node[evid] (e2) at (6.2, 0.8) {\textit{``Bimodal age dist.}\\
\textit{with nodular sclerosis}\\
\textit{most common\ldots''}};

\draw[arr, orange!70, dashed] (n4) -- (e1);
\draw[arr, orange!70, dashed] (n3) -- (e2);

\node[qnode, fill=green!10] (ans) at (6.2, -0.4)
  {\textbf{Hodgkin lymphoma}};
\draw[arr] (e1.east) -- ++(0.7, 0) |- (ans.east);
\draw[arr] (e2) -- (ans);

\node[matched, minimum size=0.25cm] at (-2.0, -1.3) {};
\node[lbl, right] at (-1.78, -1.3) {matched};
\node[reached, minimum size=0.25cm] at (-0.2, -1.3) {};
\node[lbl, right] at (0.02, -1.3) {reached};
\node[outcome, minimum size=0.25cm] at (1.6, -1.3) {};
\node[lbl, right] at (1.82, -1.3) {outcome};
\node[unreach, minimum size=0.25cm] at (3.4, -1.3) {};
\node[lbl, right] at (3.62, -1.3) {not reached};

\end{tikzpicture}}
\caption{Graph-guided evidence retrieval on a Hodgkin lymphoma query.
The query embeds and matches variable nodes (green). Bounded BFS traverses
causal edges (blue arrows) to reach an outcome node (double circle).
Evidence chunks attached to traversed edges are collected and passed to
the LLM, which produces the correct diagnosis. Grey nodes are
not connected to matched nodes.}
\label{fig:retrieval}
\end{figure}
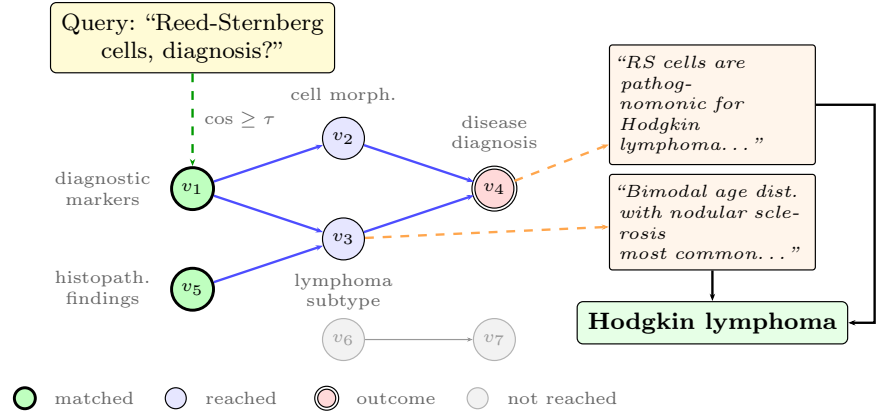

At query time, the graph acts as a \emph{structured retrieval filter}
(Figure~\ref{fig:retrieval}) in three steps: (1)~\emph{per-metric
node matching} compares the query embedding to variable embeddings
under a per-metric cosine gate that caps matches per metric; (2)~a
\emph{bounded BFS} ($h_{\max}{=}4$) from the
matched set traverses
directed causal edges, collecting reachable nodes and connecting
edges; (3)~\emph{outcome cones}, the reverse-reachable sets of
outcome/terminal nodes, are ranked by matched-node convergence and
merged by metric key, and the top $C_{\max}{=}6$ cones define the
final context. Evidence texts attached to traversed edges (via
$\phi$) are deduplicated and passed to the answer LLM; no graph
structure, node names, or edge triples are included. Formal
definitions (Defs.~\ref{def:matching}--\ref{def:cone}) and default
parameter values are in Appendix~\ref{app:algorithm}.

\section{Experimental setup}
\label{sec:experiments}

\subsection{Benchmarks}

\noindent\textbf{GraphRAG-Bench:}
We evaluate on the Medical (2,062 questions) and Novel (2,010
questions) subsets of GraphRAG-Bench. Medical tests structured
domain reasoning, while Novel provides a literary boundary case where
entity-relation structure differs substantially from the intended
setting. See details in Appendix~\ref{app:benchmarks}.
\textbf{\epilepsyqa{}:}
We curate a 375-question epilepsy benchmark from a clinical corpus
derived from two open-access textbooks. Candidate Q--A pairs were
LLM-generated and then reviewed by two board-certified neurologists,
who corrected reference answers and removed inaccurate or off-topic
items. All retained questions target epilepsy specifically. The final
benchmark spans six question types; see details in Appendix~\ref{app:benchmarks}.

\subsection{Baselines}

All systems use the same LLM (\texttt{gpt-5.4-mini}) for generation
and the same embedding model (\texttt{text-embedding-3-small}) for
retrieval, to ensure fair comparison. \sg{} additionally uses
\texttt{gpt-5.4-mini} for all graph-construction steps. The
LLM-as-judge~\citep{zheng2024judging} is held constant across systems
at \textbf{GLM-4.7}~\citep{glm47}, accessed through a LiteLLM
proxy~\citep{litellm} fronting AWS Bedrock with multi-region
routing. Graph baselines are \textbf{\msgr{}}~\citep{edge2024local},
\textbf{\lgr{}}~\citep{guo2024lightrag}, and \textbf{\fgr{}}~\citep{circlemind2024fast}
(Section~\ref{sec:related}). \textbf{\nrag{}} retrieves the top-5 chunks
(1{,}200 words, 200-word overlap) by embedding similarity without a graph.
\textbf{In-context} passes the full corpus in the answer LLM's context window with no retrieval.
\textbf{\sg{} (ours)} is specified in Section~\ref{sec:method}; ablations appear in Section~\ref{sec:ablation}.

\subsection{Evaluation metrics}

Following GraphRAG-Bench, we report LLM-judge metrics per question
type: \textbf{Answer correctness} (weighted combination of atomic-fact
$F_1$ at 0.75 and embedding similarity at 0.25);
\textbf{ROUGE-L} (lexical overlap with ground truth);
\textbf{Coverage} (fraction of ground-truth facts attributed); and \textbf{Faithfulness}
(fraction of answer statements supported by retrieved context).

\subsection{Graph scaling tiers and reproducibility}

We construct up to four \sg{} variants per corpus by capping the
specification's metric/dimension counts: medical yields
XS~(333~nodes)/S~(902)/M~(1{,}064)/L~(1{,}583), and epilepsy
XS~(218)/S~(427)/M~(468)/L~(624) under the same cap sequence
(node counts differ across corpora because the same cap produces
different graphs on corpora of different size and density). Node counts for
\sg{} are typed variables; for baselines they are extracted entities,
so edge counts are not directly comparable across paradigms. All
experiments use isolated per-framework virtual environments,
temperature~0, and report means $\pm$ standard deviations over five
independent runs (run~0 builds the graph; subsequent runs reuse it).

\section{Results}
\label{sec:results}

\subsection{Graph compactification}

The Nodes column of Table~\ref{tab:main-results} shows the
compactness gap: \sg{} achieves a 3--17$\times$ node reduction over
entity-relation baselines on medical. The smallest tier (XS,
333~nodes) is ${\sim}$17$\times$ smaller than \fgr{}, and even the
largest tier (L, 1{,}583~nodes) uses 3--4$\times$ fewer nodes.
Edge counts and wiring density are in
Table~\ref{tab:graph-sizes} (Appendix~\ref{app:stratified}).

Beyond node count, compactness has practical consequences. At
333--1{,}064 nodes, the full variable vocabulary fits in a single
editor pane (the medical
schema is 173~metrics; Appendix~\ref{app:algorithm}): a domain expert
can read every node, correct a mislabeled edge, and have the fix
propagate deterministically through graph build and retrieval.
Entity-relation graphs of 5k+ nodes offer no such affordance. The
compactness also constrains query-time cost: each \sg{} query makes a
single LLM call (answer generation from retrieved evidence), versus 2
for \lgr{}/\fgr{} and ${\sim}$196 for \msgr{} global. Per-query
latency ranges from 3.2\,s (XS) to 12.6\,s (L), comparable to
\fgr{} (2.7\,s) at the XS tier; full build and query cost details
are in Appendix~\ref{app:cost}.

\subsection{Overall answer quality}

On medical (Table~\ref{tab:main-results}; 5-run means $\pm$ std),
\msgr{} global search leads on coverage (0.649) but has the lowest
correctness (0.573) and ROUGE (0.139): community-level abstractions
lose factual specificity. \sg{}~M is statistically indistinguishable from \lgr{} on correctness
(0.679 vs.\ 0.673; $p{=}.27$, Appendix~\ref{app:significance}) and
trails \fgr{} (0.690) by 1.1\,pp, while using ${\sim}$5$\times$
fewer nodes. Treating answer quality and
graph size jointly, the smaller \sg{} tiers (XS, S, M) lie on the
medical (correctness, coverage, nodes) Pareto frontier; the L tier
is dominated by M, signalling diminishing returns past 1k nodes.
\fgr{} achieves the highest medical correctness ($0.690$) but at
$5\times$ the node count; \sg{}~M sacrifices 1.1\,pp in correctness
to gain $+9.1$\,pp in coverage with a $5\times$ smaller graph
(Figure~\ref{fig:pareto}, Appendix~\ref{app:pareto}). Paired-bootstrap
significance for every headline pair, plus the cross-stratum
hop-heavy/lookup-heavy slice analysis on epilepsy, is reported in
Appendix~\ref{app:significance}; cross-family judge robustness
(Kimi-K2.5) is in Appendix~\ref{app:cross-judge}.

\begin{table}[!ht]
\caption{Answer quality and graph size on GraphRAG-Bench medical
(2{,}062 questions, 5 runs, GLM-4.7 judge). \sg{} tiers use
3--17$\times$ fewer nodes than entity-relation baselines while landing
within 2.6\,pp of the best system on answer correctness.
\textbf{Bold} marks the best value per column;
\uline{underline} marks the best \sg{} value, which in every case is
Pareto-optimal (no baseline achieves both higher quality and fewer
nodes). ROUGE averaged over Fact Retrieval and Complex Reasoning;
Coverage over Contextual Summarization and Creative Generation.
Faithfulness (Creative Generation only) is reported in
Appendix~\ref{app:stratified}.}
\label{tab:main-results}
\centering
\begin{tabular}{lrccc}
\toprule
Framework & Nodes $\downarrow$ & Ans.\ Corr.\ $\uparrow$ & ROUGE $\uparrow$ & Coverage $\uparrow$ \\
\midrule
\nrag{}            & \textemdash & .608{\tiny$\pm$.005} & .317{\tiny$\pm$.004} & .547{\tiny$\pm$.009} \\
\midrule
\msgr{} local      & 5{,}130 & .654{\tiny$\pm$.001} & .201{\tiny$\pm$.001} & .632{\tiny$\pm$.006} \\
\msgr{} global     & 5{,}130 & .573{\tiny$\pm$.003} & .139{\tiny$\pm$.001} & \textbf{.649}{\tiny$\pm$.007} \\
\lgr{}             & 4{,}867 & .673{\tiny$\pm$.001} & .343{\tiny$\pm$.000} & .466{\tiny$\pm$.004} \\
\fgr{}             & 5{,}807 & \textbf{.690}{\tiny$\pm$.001} & \textbf{.373}{\tiny$\pm$.001} & .525{\tiny$\pm$.005} \\
\specialrule{.08em}{.3em}{.3em}
\sg{} (ours) L     & 1{,}583 & .669{\tiny$\pm$.003} & .337{\tiny$\pm$.003} & .601{\tiny$\pm$.004} \\
\sg{} (ours) M     & 1{,}064 & \uline{.679}{\tiny$\pm$.001} & \uline{.343}{\tiny$\pm$.001} & \uline{.616}{\tiny$\pm$.006} \\
\sg{} (ours) S     & 902     & .668{\tiny$\pm$.002} & .339{\tiny$\pm$.001} & .589{\tiny$\pm$.009} \\
\sg{} (ours) XS    & \textbf{333} & .664{\tiny$\pm$.003} & .336{\tiny$\pm$.001} & .570{\tiny$\pm$.007} \\
\bottomrule
\end{tabular}
\end{table}

Four qualitative case studies on medical Complex Reasoning (multi-hop
causal chains where \sg{} beats \fgr{} by $>$0.3 correctness) appear in
Appendix~\ref{app:qualitative}.

On the literary novel benchmark (Appendix~\ref{app:novel},
Table~\ref{tab:novel-results}), entity-relation systems have an
advantage: \fgr{} leads correctness (0.596), \sg{} trails \nrag{}
(0.524 vs.\ 0.544) but is comparable on coverage (0.462 vs.\ 0.488);
the In-context ablation (Table~\ref{tab:ablation}) reaches 0.591
here, the only benchmark where removing the graph
helps.\footnote{Only the S tier is reported on novel because the
medical and epilepsy scaling curves (Section~\ref{sec:scaling}) show
that tier choice has minimal effect on answer correctness ($\leq
0.015$\,pp spread); given the additional build cost for a corpus of
20~novels and the clear domain mismatch, we report a single
representative tier rather than all four.}

On \textbf{\epilepsyqa{}}
(Table~\ref{tab:epilepsy-results}), the structural separation is stark:
all four \sg{} tiers ($0.604$--$0.611$) and \msgr{}~local ($0.606$)
cluster at the top, well above \fgr{} ($0.569$), \nrag{} ($0.397$),
and \lgr{} ($0.334$). The two methods that beat flat entity-relation
retrieval are the two that impose higher-level
organization, namely \msgr{} (hierarchical community summaries) and
\sg{} (typed causal schema).
The wider gap relative to medical is consistent with the benchmark's
question distribution: 80\% of \epilepsyqa{} questions require
multi-step reasoning (Mechanistic, Multi-hop, Counterfactual, Clinical
Reasoning), while Fact Retrieval the category where flat entity
co-occurrence suffices accounts for only 1.6\%, compared with 53\%
on medical. The underlying corpus (two focused clinical textbooks) is
also more tightly covered by the typed schema than the broader medical
corpus, so the graph captures a larger share of the relationships the
questions test.
\sg{} achieves this clustering with graphs $7$--$20\times$
smaller than \msgr{}'s ($218$--$624$ vs.\ $4{,}335$ nodes;
Table~\ref{tab:epilepsy-results}). The structural separation is
robust to a cross-family second judge (Kimi-K2.5; both \sg{} and
\msgr{}~local significantly outperform every flat entity-relation
baseline under both judges, $q{<}.001$;
Appendix~\ref{app:cross-judge}). On secondary metrics, \sg{}'s
top tiers lead aggregate ROUGE (0.206 vs.\ 0.201 \fgr{}, 0.170
\msgr{}~local) and \sg{}~XS leads Fact Retrieval ($+10.7$\,pp over
\msgr{}~local; Appendix~\ref{app:epilepsy-stratified}).
The separation is structural rather than graph-size driven:
\msgr{} adds Leiden communities and multi-resolution summaries on
top of its entity-relation graph, so the two systems that beat flat
entity-relation retrieval are the two that impose higher-level
organization (\fgr{}'s entity-relation graph trails by 3.7\,pp at
comparable size to \msgr{}'s). Per-question-type breakdowns are in
Appendices~\ref{app:stratified} and~\ref{app:epilepsy-stratified}.

\begin{table}[!ht]
\caption{Answer quality and graph size on \epilepsyqa{} (375
neurologist-validated questions, 5 runs, GLM-4.7 judge). All four
\sg{} tiers (0.604--0.611) and \msgr{}~local (0.606) cluster at the
top, significantly above flat entity-relation methods, while using
7--20$\times$ fewer nodes. Emphasis conventions follow
Table~\ref{tab:main-results}. ROUGE averaged over all non-synthesis
types; Coverage over Synthesis only.}
\label{tab:epilepsy-results}
\centering
\begin{tabular}{lrccc}
\toprule
Framework & Nodes $\downarrow$ & Ans.\ Corr.\ $\uparrow$ & ROUGE $\uparrow$ & Coverage $\uparrow$ \\
\midrule
\nrag{}            & \textemdash & .397{\tiny$\pm$.006} & .124{\tiny$\pm$.002} & .212{\tiny$\pm$.007} \\
\midrule
\msgr{} local      & 4{,}335 & .606{\tiny$\pm$.005} & .170{\tiny$\pm$.001} & .370{\tiny$\pm$.010} \\
\msgr{} global     & 4{,}335 & .562{\tiny$\pm$.005} & .140{\tiny$\pm$.001} & \textbf{.464}{\tiny$\pm$.005} \\
\lgr{}             & 4{,}148 & .334{\tiny$\pm$.002} & .095{\tiny$\pm$.000} & .144{\tiny$\pm$.004} \\
\fgr{}             & 5{,}227 & .569{\tiny$\pm$.005} & .201{\tiny$\pm$.003} & .311{\tiny$\pm$.010} \\
\specialrule{.08em}{.3em}{.3em}
\sg{} (ours) L     & 624     & \textbf{.611}{\tiny$\pm$.006} & \textbf{.206}{\tiny$\pm$.002} & .422{\tiny$\pm$.007} \\
\sg{} (ours) M     & 468     & .604{\tiny$\pm$.004} & .205{\tiny$\pm$.001} & \uline{.425}{\tiny$\pm$.009} \\
\sg{} (ours) S     & 427     & .604{\tiny$\pm$.006} & .206{\tiny$\pm$.002} & .406{\tiny$\pm$.012} \\
\sg{} (ours) XS    & \textbf{218} & .607{\tiny$\pm$.005} & .201{\tiny$\pm$.001} & .397{\tiny$\pm$.021} \\
\bottomrule
\end{tabular}
\end{table}

\subsection{Scaling analysis}
\label{sec:scaling}

Quality vs.\ graph size shows \emph{diminishing returns past M on
medical}: XS$\to$S$\to$M improves monotonically, but expanding to L
yields no net gain on the headline metric and slight regressions on
Complex Reasoning ($-$1.3\,pp) and Creative Generation faithfulness
($-$2.8\,pp), consistent with schema saturation. M and S have
similar edge counts (8{,}455 vs.\ 8{,}846) despite differing node
counts because both tiers' metric-level caps produced comparable
Tier-1 edge templates on this corpus; the node difference reflects
variable instantiation, not edge discovery. On epilepsy the curve is
flatter still, with all four tiers within 0.010 answer correctness;
even the XS tier (218~nodes) captures the domain's causal structure.
Full tier-by-tier breakdowns are in
Appendices~\ref{app:stratified} (medical)
and~\ref{app:epilepsy-stratified} (epilepsy).

\section{Ablation study}
\label{sec:ablation}

Two ablations on the S tier isolate the role of the graph
(Table~\ref{tab:ablation}).
\emph{Nodes only} runs the same BFS over the same graph but feeds the
LLM only the matched node names and edge triples. \emph{In-context} removes the graph
entirely and passes the full corpus to the LLM.

\begin{table}[!ht]
\caption{Ablation: medical answer correctness (5-run mean $\pm$ sample std
across runs, computed from count-weighted answer correctness in evaluation exports).
\sg{} (default) provides only graph-selected evidence to the answer LLM; the
per-question-type breakdown (including ablation rows) is in
Appendix~\ref{app:stratified}, Table~\ref{tab:stratified}.}
\label{tab:ablation}
\centering
\footnotesize
\begin{tabular}{lc}
\toprule
Variant            & Medical $\uparrow$ \\
\midrule

\sg{} (ours) S (default)  & \textbf{.668}{\tiny$\pm$.002} \\
\midrule
Nodes only         & .480{\tiny$\pm$.006} \\
In-context         & .505{\tiny$\pm$.003} \\
\bottomrule
\end{tabular}

\vspace{3pt}
{\scriptsize\textit{Nodes only}: same BFS, LLM receives node names
and edge triples instead of evidence passages. \textit{In-context}:
no retrieval; LLM receives the full corpus.}
\end{table}

\paragraph{The graph is a retrieval filter, not a reasoning substrate.}
Default \sg{} feeds only evidence passages to the answer LLM and
reaches 0.668 on medical, within 2.2\,pp of \fgr{} (0.690) and
statistically indistinguishable from \lgr{} (0.673). \emph{Nodes
only} inverts this and drops to 0.480 ($-18.8$\,pp), confirming that
the LLM relies on retrieved text rather than graph structure itself.
The gap to \nrag{} (0.608) is $+6.0$\,pp ($p{<}.001$, paired
bootstrap; Appendix~\ref{app:significance}). Both \nrag{} and \sg{}
feed only text to the LLM, so the graph's role is to \emph{select}
evidence via BFS, a structured filter distinct from top-$k$ search
and PageRank. Removing retrieval entirely (\emph{In-context}) costs
$-16.3$\,pp on medical and $-15.9$\,pp on epilepsy, but reaches
0.591 on novel, consistent with the finding that the causal graph's
retrieval advantage is strongest on structured domains.

\section{Limitations}
\label{sec:limitations}

\noindent\textbf{Graph correctness scope.}
Neurologists validated the \epilepsyqa{} questions
(Table~\ref{tab:epilepsy-results}), and the compact typed graph is
small enough to inspect in a single editor pane, a deliberate design
goal. Every causal edge stores mechanism text and chunk-level
provenance, so a domain expert can verify or correct individual claims
without re-running the pipeline. A systematic edge-by-edge audit was
beyond the scope of this study; we view it as a natural next step
enabled precisely by the graph's compactness.
\textbf{Precision--coverage trade-off.}
Schema-constrained construction deliberately favors precision over
broad lexical coverage, which is the appropriate trade-off for
fact-heavy, causally structured domains. On summarization-heavy tasks
(Contextual Summarization, Creative Generation, epilepsy Synthesis),
\msgr{} global's community summaries achieve higher aggregate
coverage.
\textbf{Domain scope.}
\sg{} is designed for corpora with discoverable hierarchical causal
structure (e.g., medical and scientific text). On literary narrative,
where such structure is largely absent, the causal graph may offer
less benefit, as the novel benchmark suggests. This is consistent
with the paper's central thesis that graph topology should match the
domain's knowledge structure rather than default to a single paradigm.

\section{Conclusion}
\label{sec:conclusion}

We presented \sg{}, a system that constructs schema-constrained
hierarchical causal graphs via an automated domain-specification
pipeline and uses them as a graph-guided retrieval filter. The
headline finding is \emph{compactness without sacrifice}:
schema-constrained causal graphs match entity-relation GraphRAG with
3--20$\times$ fewer nodes, producing graphs small enough for a domain
expert to audit, correct, and extend. Ablations attribute the gain to
the causal graph itself, which serves as a \emph{structured retrieval
filter}. The second finding is structural: on specialized domains with
discoverable hierarchical causal structure (epilepsy), only methods
that impose higher-level organization (\msgr{}'s community summaries
and \sg{}'s typed causal schema) outperform flat entity-relation
retrieval; on literary narrative, where such structure is largely
absent, entity-relation graphs lead. Because the pipeline produces
inspectable artifacts (YAML schema, typed nodes, per-edge mechanism
text) over a graph that fits in a single editor pane, our fully
automated numbers are a \emph{zero-shot lower bound} on what the
approach can achieve with expert-in-the-loop curation.
The broader design lesson is to ask \emph{what} to put in the
graph and \emph{how} to use it, not only how many nodes to extract.

\bibliographystyle{plainnat}
\bibliography{references}

@inproceedings{graphragbench,
  title={When to Use Graphs in {RAG}: A Comprehensive Analysis for Graph Retrieval-Augmented Generation},
  author={Xiang, Zhishang and Wu, Chuanjie and Zhang, Qinggang and Chen, Shengyuan and Hong, Zijin and Huang, Xiao and Su, Jinsong},
  booktitle={International Conference on Learning Representations},
  year={2026},
  note={Introduces the GraphRAG-Bench benchmark. \url{https://github.com/GraphRAG-Bench/GraphRAG-Benchmark}}
}

@article{edge2024local,
  title={From Local to Global: A {GraphRAG} Approach to Query-Focused Summarization},
  author={Edge, Darren and Trinh, Ha and Cheng, Newman and Bradley, Joshua and Chao, Alex and Mody, Apurva and Truitt, Steven and Metropolitansky, Dasha and Ness, Robert Osazuwa and Larson, Jonathan},
  journal={arXiv preprint arXiv:2404.16130},
  year={2024}
}

@article{guo2024lightrag,
  title={{LightRAG}: Simple and Fast Retrieval-Augmented Generation},
  author={Guo, Zirui and Xia, Lianghao and Yu, Yanhua and Ao, Tu and Huang, Chao},
  journal={arXiv preprint arXiv:2410.05779},
  year={2024}
}

@article{circlemind2024fast,
  title={{Fast GraphRAG}: {RAG} that Intelligently Adapts to Your Use Case, Data, and Queries},
  author={{Circlemind AI}},
  journal={GitHub repository},
  year={2024},
  note={\url{https://github.com/circlemind-ai/fast-graphrag}}
}

@article{lewis2020rag,
  title={Retrieval-Augmented Generation for Knowledge-Intensive {NLP} Tasks},
  author={Lewis, Patrick and Perez, Ethan and Piktus, Aleksandra and Petroni, Fabio and Karpukhin, Vladimir and Goyal, Naman and K{\"u}ttler, Heinrich and Lewis, Mike and Yih, Wen-tau and Rockt{\"a}schel, Tim and Riedel, Sebastian and Kiela, Douwe},
  journal={Advances in Neural Information Processing Systems},
  volume={33},
  pages={9459--9474},
  year={2020}
}

@article{gao2024rag,
  title={Retrieval-Augmented Generation for Large Language Models: A Survey},
  author={Gao, Yunfan and Xiong, Yun and Gao, Xinyu and Jia, Kangxiang and Pan, Jinliu and Bi, Yuxi and Dai, Yi and Sun, Jiawei and Guo, Qianyu and Wang, Meng and Wang, Haofen},
  journal={arXiv preprint arXiv:2312.10997},
  year={2024}
}

@inproceedings{karpukhin2020dense,
  title={Dense Passage Retrieval for Open-Domain Question Answering},
  author={Karpukhin, Vladimir and Oguz, Barlas and Min, Sewon and Lewis, Patrick and Wu, Ledell and Edunov, Sergey and Chen, Danqi and Yih, Wen-tau},
  booktitle={Proceedings of the 2020 Conference on Empirical Methods in Natural Language Processing (EMNLP)},
  pages={6769--6781},
  year={2020}
}

@inproceedings{guu2020realm,
  title={{REALM}: Retrieval-Augmented Language Model Pre-Training},
  author={Guu, Kelvin and Lee, Kenton and Tung, Zora and Pasupat, Panupong and Chang, Ming-Wei},
  booktitle={Proceedings of the 37th International Conference on Machine Learning},
  pages={3929--3938},
  year={2020}
}

@inproceedings{izacard2021leveraging,
  title={Leveraging Passage Retrieval with Generative Models for Open Domain Question Answering},
  author={Izacard, Gautier and Grave, Edouard},
  booktitle={Proceedings of the 16th Conference of the European Chapter of the Association for Computational Linguistics: Main Volume},
  pages={874--880},
  year={2021}
}

@inproceedings{jiang2023active,
  title={Active Retrieval Augmented Generation},
  author={Jiang, Zhengbao and Xu, Frank F. and Gao, Luyu and Sun, Zhiqing and Liu, Qian and Dwivedi-Yu, Jane and Yang, Yiming and Callan, Jamie and Neubig, Graham},
  booktitle={Proceedings of the 2023 Conference on Empirical Methods in Natural Language Processing},
  pages={7969--7992},
  year={2023}
}

@inproceedings{trivedi2023interleaving,
  title={Interleaving Retrieval with Chain-of-Thought Reasoning for Knowledge-Intensive Multi-Step Questions},
  author={Trivedi, Harsh and Balasubramanian, Niranjan and Khot, Tushar and Sabharwal, Ashish},
  booktitle={Proceedings of the 61st Annual Meeting of the Association for Computational Linguistics (Volume 1: Long Papers)},
  pages={10014--10037},
  year={2023}
}

@inproceedings{asai2024selfrag,
  title={Self-{RAG}: Learning to Retrieve, Generate, and Critique through Self-Reflection},
  author={Asai, Akari and Wu, Zeqiu and Wang, Yizhong and Sil, Avirup and Hajishirzi, Hannaneh},
  booktitle={International Conference on Learning Representations},
  year={2024}
}

@article{pan2024kgllm,
  title={Unifying Large Language Models and Knowledge Graphs: A Roadmap},
  author={Pan, Shirui and Luo, Linhao and Wang, Yufei and Chen, Chen and Wang, Jiapu and Wu, Xindong},
  journal={IEEE Transactions on Knowledge and Data Engineering},
  volume={36},
  number={7},
  pages={3580--3599},
  year={2024},
  doi={10.1109/TKDE.2024.3352100}
}

@inproceedings{baek2023kaping,
  title={Knowledge-Augmented Language Model Prompting for Zero-Shot Knowledge Graph Question Answering},
  author={Baek, Jinheon and Aji, Alham Fikri and Saffari, Amir},
  booktitle={Proceedings of the 1st Workshop on Natural Language Reasoning and Structured Explanations (NLRSE)},
  pages={78--106},
  year={2023}
}

@inproceedings{he2024gretriever,
  title={{G-Retriever}: Retrieval-Augmented Generation for Textual Graph Understanding and Question Answering},
  author={He, Xiaoxin and Tian, Yijun and Sun, Yifei and Chawla, Nitesh V. and Laurent, Thomas and LeCun, Yann and Bresson, Xavier and Hooi, Bryan},
  booktitle={Advances in Neural Information Processing Systems},
  year={2024}
}

@inproceedings{sun2019pullnet,
  title={{PullNet}: Open Domain Question Answering with Iterative Retrieval on Knowledge Bases and Text},
  author={Sun, Haitian and Bedrax-Weiss, Tania and Cohen, William W.},
  booktitle={Proceedings of the 2019 Conference on Empirical Methods in Natural Language Processing and the 9th International Joint Conference on Natural Language Processing (EMNLP-IJCNLP)},
  pages={2380--2390},
  year={2019}
}

@inproceedings{oguz2022unikqa,
  title={{UniK-QA}: Unified Representations of Structured and Unstructured Knowledge for Open-Domain Question Answering},
  author={Oguz, Barlas and Chen, Xilun and Karpukhin, Vladimir and Peshterliev, Stan and Okhonko, Dmytro and Schlichtkrull, Michael and Gupta, Sonal and Mehdad, Yashar and Yih, Scott},
  booktitle={Findings of the Association for Computational Linguistics: NAACL 2022},
  pages={1535--1546},
  year={2022}
}

@article{glymour2019review,
  title={Review of Causal Discovery Methods Based on Graphical Models},
  author={Glymour, Clark and Zhang, Kun and Spirtes, Peter},
  journal={Frontiers in Genetics},
  volume={10},
  pages={524},
  year={2019}
}

@article{kiciman2024causal,
  title={Causal Reasoning and Large Language Models: Opening a New Frontier for Causality},
  author={K{\i}c{\i}man, Emre and Ness, Robert Osazuwa and Sharma, Amit and Tan, Chenhao},
  journal={Transactions on Machine Learning Research},
  year={2024},
  note={Selected for presentation at ICLR 2025}
}

@article{jiralerspong2024efficient,
  title={Efficient Causal Graph Discovery Using Large Language Models},
  author={Jiralerspong, Thomas and Chen, Xiaoyin and More, Yash and Shah, Vedant and Bengio, Yoshua},
  journal={arXiv preprint arXiv:2402.01207},
  year={2024}
}

@inproceedings{wang2025causalrag,
  title={{CausalRAG}: Integrating Causal Graphs into Retrieval-Augmented Generation},
  author={Wang, Nengbo and Han, Xiaotian and Singh, Jagdip and Ma, Jing and Chaudhary, Vipin},
  booktitle={Findings of the Association for Computational Linguistics: ACL 2025},
  pages={22680--22693},
  year={2025}
}

@inproceedings{jin2023cladder,
  title={{CLadder}: Assessing Causal Reasoning in Language Models},
  author={Jin, Zhijing and Chen, Yuen and Leeb, Felix and Gresele, Luigi and Kamal, Ojasv and Lyu, Zhiheng and Blin, Kevin and Gonzalez Adauto, Fernando and Kleiman-Weiner, Max and Sachan, Mrinmaya and Sch{\"o}lkopf, Bernhard},
  booktitle={Advances in Neural Information Processing Systems},
  volume={36},
  pages={31038--31065},
  year={2023}
}

@inproceedings{pal2022medmcqa,
  title={{MedMCQA}: A Large-scale Multi-Subject Multi-Choice Dataset for Medical Domain Question Answering},
  author={Pal, Ankit and Umapathi, Logesh Kumar and Sankarasubbu, Malaikannan},
  booktitle={Proceedings of the Conference on Health, Inference, and Learning},
  pages={248--260},
  year={2022}
}

@article{singhal2023large,
  title={Large Language Models Encode Clinical Knowledge},
  author={Singhal, Karan and Azizi, Shekoofeh and Tu, Tao and Mahdavi, S. Sara and Wei, Jason and Chung, Hyung Won and Scales, Nathan and Tanwani, Ajay and Cole-Lewis, Heather and Pfohl, Stephen and others},
  journal={Nature},
  volume={620},
  pages={172--180},
  year={2023}
}

@book{bromfield2006epilepsy,
  title={An Introduction to Epilepsy},
  editor={Bromfield, Edward B. and Cavazos, Jos{\'e} E. and Sirven, Joseph I.},
  year={2006},
  publisher={American Epilepsy Society},
  address={West Hartford, CT},
  note={NCBI Bookshelf ID: NBK2508. Creative Commons Attribution. Available at \url{https://www.ncbi.nlm.nih.gov/books/NBK2508/}}
}

@book{czuczwar2022epilepsy,
  title={Epilepsy},
  editor={Czuczwar, Stanis{\l}aw J},
  year={2022},
  publisher={Exon Publications},
  address={Brisbane, Australia},
  isbn={978-0-6453320-4-9},
  doi={10.36255/exon-publications-epilepsy},
  note={NCBI Bookshelf ID: NBK580617. Electronic chapters available under Creative Commons Attribution--NonCommercial 4.0 at \url{https://www.ncbi.nlm.nih.gov/books/NBK580617/}}
}

@inproceedings{zheng2024judging,
  title={Judging {LLM}-as-a-{J}udge with {MT}-{B}ench and {C}hatbot {A}rena},
  author={Zheng, Lianmin and Chiang, Wei-Lin and Sheng, Ying and Zhuang, Siyuan and Wu, Zhanghao and Zhuang, Yonghao and Lin, Zi and Li, Zhuohan and Li, Dacheng and Xing, Eric P. and Zhang, Hao and Gonzalez, Joseph E. and Stoica, Ion},
  booktitle={Advances in Neural Information Processing Systems},
  volume={36},
  year={2023}
}

@misc{glm47,
  title  = {{GLM-4.7}: Advancing the Coding Capability},
  author = {{Z.ai}},
  year   = {2025},
  note   = {\url{https://z.ai/blog/glm-4.7}}
}

@misc{litellm,
  title  = {{LiteLLM}: Unified {API} for {LLM} Providers},
  author = {{BerriAI}},
  year   = {2023},
  note   = {Software library, first released 2023. \url{https://github.com/BerriAI/litellm}}
}

\appendix
\section{Method details}
\label{app:algorithm}

This appendix collects the algorithmic specification, the
retrieval-stage definitions (per-metric node matching, bounded BFS,
outcome cone), and the derivation for
Remark~\ref{rem:search-space}, all referenced from
Section~\ref{sec:method}.

\paragraph{Derivation for Remark~\ref{rem:search-space}.}
Tier~1 evaluates at most $|\Metrics|^2 - |\Metrics|$ ordered metric
pairs (after excluding self-loops and terminal sources). Tier~2
evaluates node pairs only for templates in $\Edges_m$: each template
$(m_s, m_t)$ contributes $|\Vars_{m_s}| \cdot |\Vars_{m_t}| \leq
\bar{k}^2$ candidates. Summing over templates,
\[
\text{Total calls} \leq \underbrace{|\Metrics|(|\Metrics|-1)}_{\text{Tier 1}}
+ \underbrace{\sum_{(m_s,m_t) \in \Edges_m} |\Vars_{m_s}| \cdot
|\Vars_{m_t}|}_{\text{Tier 2}}
\leq |\Metrics|^2 + |\Edges_m| \cdot \bar{k}^2.
\]
For exhaustive pairwise evaluation, all $|\Vars|^2$ pairs must be
considered. The two-tier bound never exceeds this: in the worst case
where every metric pair yields a template
($|\Edges_m| = |\Metrics|^2$), the Tier~2 sum becomes
$\sum_{m_s, m_t} |\Vars_{m_s}| \cdot |\Vars_{m_t}|
= \bigl(\sum_m |\Vars_m|\bigr)^2 = |\Vars|^2$,
recovering the naive bound. In practice $|\Edges_m| \ll |\Metrics|^2$
since Tier~1 filters most metric pairs, yielding a strict
reduction.\hfill$\square$

\FloatBarrier
\paragraph{Inspectability: a representative schema excerpt.}
The fixed metric/dimension vocabulary (FIXED in
Figure~\ref{fig:domspec}) is small enough to print and read in a
sitting. Figure~\ref{fig:schema-excerpt} shows a 28-line excerpt
from the medical schema's metrics block: each metric is named with
its dimension signature, has a one-paragraph mechanism description,
and carries a graph role. The full medical schema is 173 metrics
across 9 categories; the full epilepsy schema is 91 metrics across
8 categories. Both fit in a single text editor pane and are
human-editable; corrections propagate deterministically through
graph build (Stage~2) and retrieval (Stage~3).

\begin{figure}[!ht]
\caption{Excerpt from the medical \texttt{graph\_spec.yaml} (3 of
173 metrics shown). The vocabulary is closed at construction time:
new entities and variables are admitted only if they fit an
existing metric template, which is what bounds graph size.}
\label{fig:schema-excerpt}
\begin{lstlisting}[language=,basicstyle=\ttfamily\scriptsize,
                   frame=single,framerule=0.4pt,
                   xleftmargin=0pt,xrightmargin=0pt,
                   columns=fullflexible,keepspaces=true]
metrics:
- name: DiagnosticApproach(disease, diagnostic_modality, biomarker)
  description: Methods used to detect and confirm disease, including
    imaging modalities, biopsies, blood tests, urine tests, biomarker
    panels, pathology review, genetic testing, flow cytometry, and
    endoscopic procedures. Driven by symptoms, risk factors, anatomy
    involved, and prior test results; influences staging accuracy,
    treatment selection, and surveillance planning.
  attributes:
    category: diagnostics
    graph_role: outcome

- name: AnatomicalInvolvement(disease, anatomy, outcome_horizon)
  description: Body sites and tissue structures involved by the
    disease over time, including primary sites, tissue layers,
    nerves, lymph nodes, marrow, and metastatic organs. Shaped by
    local invasion and metastatic spread; influences symptoms,
    staging, treatment options, and prognosis.
  attributes:
    category: anatomy
    graph_role: outcome

- name: StagingClassification(disease, stage, subtype)
  description: Clinical or pathologic severity classification
    reflecting disease extent, grade, phase, molecular subgroup, or
    risk category, including TNM stage, local/regional/metastatic
    extent, leukemia phase, and low/high/very-high-risk groupings.
    Determined by diagnostic findings, invasion extent, metastatic
    burden, and biomarker status; directly drives therapy choice
    and prognosis.
  attributes:
    category: staging
    graph_role: outcome
\end{lstlisting}
\end{figure}

\begin{algorithm}[ht]
\caption{\sg{} pipeline}
\label{alg:pipeline}
\begin{algorithmic}[1]
\REQUIRE Corpus $\mathcal{D} = \{d_1, \ldots, d_n\}$, outcome config $\mathcal{O}$, query $q$
\ENSURE Answer $a$

\STATE \textbf{Stage 1: Domain Specification}
\STATE $\Dim \leftarrow \textsc{GenerateDimensions}(\mathcal{D})$
  \COMMENT{axes $(\delta_i, E_i)$; Definition~\ref{def:spec}}
\STATE $\Metrics \leftarrow \textsc{GenerateMetrics}(\mathcal{D}, \Dim)$
  \COMMENT{fixed metrics and signatures $\operatorname{sig}(m)$}
\STATE $\Vars_0 \leftarrow \textsc{GenerateVariables}(\mathcal{D}, \Dim, \Metrics)$
  \COMMENT{seed variables $v=(m,\mathbf{e})$ (extensible)}
\STATE $\Rels_0 \leftarrow \textsc{GenerateCausalRelations}(\mathcal{D}, \Dim, \Metrics, \Vars_0)$
  \COMMENT{seed edges on $\Vars_0$ (extensible)}

\STATE \textbf{Stage 2: Graph Build}
\STATE $\Edges_m \leftarrow \textsc{MetricEdges}(\Metrics, \Rels_0, \mathcal{D})$
  \COMMENT{metric-level templates}
\STATE $(\Vars, \Edges_n) \leftarrow \textsc{NodeEdges}(\Vars_0, \Edges_m, \mathcal{D})$
  \COMMENT{tier-2 instantiation; may extend $\Vars$ beyond $\Vars_0$}
\STATE $G \leftarrow \textsc{LinkEvidence}((\Vars, \Edges_n), \mathcal{D})$
  \COMMENT{attach evidence chunks}

\STATE \textbf{Stage 3: Query-Time Retrieval}
\STATE $V_q \leftarrow \textsc{MatchNodes}(q, \Vars)$
  \COMMENT{per-metric cosine gating}
\STATE $G_q \leftarrow \textsc{BFS}(G, V_q, h_{\max})$
  \COMMENT{bounded causal subgraph}
\STATE $C_q \leftarrow \textsc{CollectEvidence}(G_q)$
  \COMMENT{evidence from traversed edges}
\STATE $a \leftarrow \textsc{LLM}(q, C_q)$
\end{algorithmic}
\end{algorithm}

\FloatBarrier
\paragraph{Retrieval-stage definitions.}
The three-step retrieval procedure summarized in
Section~\ref{sec:retrieval} is defined formally as follows.

\begin{definition}[Per-Metric Node Matching]
\label{def:matching}
Given query embedding $\mathbf{q}$, pre-computed variable embeddings
$\{\mathbf{v}_i\}$, and metric assignment $\mu: \Vars \to \Metrics$,
the match set $V_q$ is constructed via a per-metric gating procedure
that prevents any single metric from dominating:
\begin{enumerate}
\item For each metric $m_j$, compute the top similarity
  $\sigma_j^* = \max_{v_i:\, \mu(v_i) = m_j}
  \cosim(\mathbf{q}, \mathbf{v}_i)$.
\item Retain metric $m_j$ iff $\sigma_j^* \geq \tau_{\min}$
  (admission floor on the metric's best variable, default $0.35$).
  Keep the top $K$ metrics by $\sigma_j^*$ (default $K = 5$).
\item For each retained metric, compute the per-variable admission
  threshold $\tau_j = \min(\tau_{\min},\; \sigma_j^* (1 - \beta))$
  with band $\beta = 0.20$. The relative term $\sigma_j^*(1-\beta)$
  enforces a band below the metric's best match; $\tau_{\min}$ then
  caps that threshold so a single very-high $\sigma_j^*$ does not
  exclude moderately-similar variables. The effective threshold is
  thus $\tau_j \leq \tau_{\min}$.
\item $V_q = \{v_i \mid \mu(v_i) = m_j,\;
  \cosim(\mathbf{q}, \mathbf{v}_i) \geq \tau_j\}$, capped at $P = 3$
  per metric, sorted by similarity.
\end{enumerate}
\end{definition}

\begin{definition}[Bounded BFS Subgraph]
\label{def:bfs}
Given causal graph $G = (\Vars, \Edges, \phi)$, match set $V_q$, and
depth bound $h_{\max}$ (default 4), the reachable subgraph
$G_q = (V_r, E_r)$ is constructed by running forward BFS from every
matched node in $V_q$ along directed causal edges for at most
$h_{\max}$ hops:
\[
V_r = \bigl\{v \in \Vars \mid \exists\, v_0 \in V_q \text{ s.t.\ }
v_0 \to \cdots \to v \text{ in } G,\;
\text{path length} \leq h_{\max}\bigr\}.
\]
$E_r$ is the set of edges in $G$ between nodes in $V_r$, and each
edge carries its linked evidence via $\phi$.
\end{definition}

\begin{definition}[Outcome Cone]
\label{def:cone}
For outcome node $o \in V_r$ (identified by
role $\in \{\text{terminal}, \text{outcome}\}$), the
\emph{outcome cone} $\mathcal{K}_o$ is the set of nodes in $G_q$ from
which $o$ is reachable via reverse BFS on $G_q$:
\[
\mathcal{K}_o = \{v \in V_r \mid v \to \cdots \to o \text{ in } G_q\}.
\]
Cones are ranked by convergence count $|\mathcal{K}_o \cap V_q|$
(how many matched query nodes reach this outcome). When multiple
outcome nodes share the same metric key (e.g., two
\texttt{PatientOutcome} variables for different entities), their cones
are merged into a single combined cone to avoid redundant causal
chains. The final set is capped at $C_{\max} = 6$ cones.
\end{definition}

\paragraph{Hyperparameters.}
The per-metric matching and BFS defaults are in
Table~\ref{tab:matching-params}. The effective per-metric threshold
$\tau_j = \min(\tau_{\min},\; \sigma_j^*(1-\beta))$ adapts to each
metric's best match quality: when
$\sigma_j^* = 0.65$, $\tau_j = \min(0.35, 0.52) = 0.35$; for a weaker
match $\sigma_j^* = 0.40$, $\tau_j = \min(0.35, 0.32) = 0.32$,
admitting variables with slightly lower similarity when the overall
signal is weaker. Defaults were determined empirically on the medical
benchmark and are likely domain-dependent; corpora with different
vocabulary density or embedding characteristics may require re-tuning,
particularly $\tau_{\min}$, $\tau_e$, and $K$.

\begin{table}[!ht]
\centering
\caption{Node matching and retrieval hyperparameters.}
\label{tab:matching-params}
\begin{tabular}{llc}
\toprule
Parameter & Description & Default \\
\midrule
$\tau_{\min}$ & Absolute minimum cosine similarity & 0.35 \\
$\beta$ & Relative threshold band & 0.20 \\
$K$ & Top-$K$ metrics retained & 5 \\
$P$ & Max matches per metric & 3 \\
$h_{\max}$ & Maximum BFS depth & 4 \\
$C_{\max}$ & Maximum outcome cones & 6 \\
$\tau_e$ & Evidence linking similarity threshold & 0.78 \\
\bottomrule
\end{tabular}
\end{table}

\FloatBarrier
\section{Evidence linking details}
\label{app:evidence}

Evidence linking (Section~\ref{sec:graph-build}) employs two
complementary strategies:
\begin{enumerate}
\item \textbf{Text matching.} For edges with document-sourced
  evidence, the first 200 characters of each evidence text are matched
  against corpus chunks via substring search. This captures exact
  textual provenance when the LLM quotes source material during graph
  building.
\item \textbf{Semantic matching.} The edge context string
  ``source $\to$ target: \textit{mechanism}'' is embedded and compared
  to pre-embedded corpus chunks via cosine similarity. Chunks with
  similarity $\geq \tau_e = 0.78$ are linked, keeping at most 2 chunks
  per edge.
\end{enumerate}
The hybrid approach ensures coverage: substring matching captures
direct textual evidence, while semantic matching links edges to
paraphrased or summarized corpus content.

\section{Benchmark details}
\label{app:benchmarks}

This appendix collects corpus and dataset details omitted from
Section~4.1 for space.

\paragraph{GraphRAG-Bench.}
We use the Medical and Novel subsets of
GraphRAG-Bench~\citep{graphragbench}. Medical contains 2{,}062
expert-authored questions across Fact Retrieval (1{,}098), Complex
Reasoning (509), Contextual Summarization (289), and Creative
Generation (166). Novel contains 2{,}010 questions across 20 novels
with the same four question types. Each item is paired with a
ground-truth answer and source corpus documents.

\paragraph{\epilepsyqa{}.}
Our epilepsy evaluation set is derived from a clinical corpus built
from two open-access textbooks~\citep{bromfield2006epilepsy,
czuczwar2022epilepsy}, one under Creative Commons Attribution and one
under Creative Commons Attribution--NonCommercial terms. We first
generated 700 candidate Q--A pairs with Claude Opus~4.6 (extended
thinking enabled), then had two board-certified neurologists from a
U.S.\ medical school independently review the full set, correcting
reference answers and removing inaccurate or off-topic items. The
final set contains 375 questions spanning six types: Mechanistic
(102), Multi-hop Causal (102), Synthesis (69), Clinical Reasoning
(57), Counterfactual (39), and Fact Retrieval (6). The release will
include benchmark questions, reference answers, neurologist
annotations, and corpus reconstruction instructions.
\section{Novel benchmark results}
\label{app:novel}

Table~\ref{tab:novel-results} reports aggregate answer quality on the
GraphRAG-Bench novel subset (2{,}010 questions, five runs). Only the
S~tier is reported here: medical and epilepsy scaling curves
(Section~\ref{sec:scaling}) show that tier choice has minimal impact
on answer correctness ($\leq 0.015$\,pp spread across four tiers),
and the novel corpus represents a known boundary case where
hierarchical causal structure is largely absent, so evaluating
additional tiers would not change the qualitative conclusion.
\fgr{} leads across all metrics; \sg{} trails \nrag{} on correctness
but is comparable on coverage. The In-context ablation
(Table~\ref{tab:ablation}) reaches 0.591 correctness on novel
compared to 0.505 on medical and 0.445 on epilepsy: removing the
graph helps only on novel, confirming that the causal graph's
advantage is domain-dependent and strongest on structured knowledge
rather than literary narrative.

\begin{table}[!ht]
\caption{Answer quality on GraphRAG-Bench novel (2{,}010 questions,
5 runs, GLM-4.7 judge). $^\dagger$\emph{In-context} feeds the
full corpus to the answer LLM with no graph retrieval (the In-context
ablation; Section~\ref{sec:ablation}); included here because it is
the only system to surpass \nrag{} on novel.}
\label{tab:novel-results}
\centering
\begin{tabular}{lcccc}
\toprule
Framework & Ans.\ Corr.\ $\uparrow$ & ROUGE $\uparrow$ & Coverage $\uparrow$ & Faithful.\ $\uparrow$ \\
\midrule
\nrag{}          & .544{\tiny$\pm$.002} & .324{\tiny$\pm$.002} & .488{\tiny$\pm$.005} & \textbf{.394}{\tiny$\pm$.023} \\
\midrule
\lgr{}           & .579{\tiny$\pm$.026} & .303{\tiny$\pm$.016} & .358{\tiny$\pm$.042} & .168{\tiny$\pm$.212} \\
\fgr{}           & \textbf{.596}{\tiny$\pm$.003} & .359{\tiny$\pm$.003} & \textbf{.517}{\tiny$\pm$.004} & .391{\tiny$\pm$.018} \\
\specialrule{.08em}{.3em}{.3em}
\sg{} (ours)     & .524{\tiny$\pm$.001} & .261{\tiny$\pm$.001} & .462{\tiny$\pm$.004} & .313{\tiny$\pm$.016} \\
In-context$^\dagger$ & .591{\tiny$\pm$.003} & \textbf{.362}{\tiny$\pm$.004} & .510{\tiny$\pm$.008} & .153{\tiny$\pm$.026} \\
\bottomrule
\end{tabular}
\end{table}

\FloatBarrier
\section{Medical detailed results}
\label{app:stratified}

Table~\ref{tab:graph-sizes} lists edge counts and wiring density for
the medical graphs (node counts appear in the main results table,
Table~\ref{tab:main-results}).
Tables~\ref{tab:stratified} and~\ref{tab:full-scaling} report
per-question-type breakdowns on GraphRAG-Bench medical (2{,}062
questions, 5 runs). Table~\ref{tab:stratified} stratifies all
baselines, all \sg{} tiers, and the two ablations from
Section~\ref{sec:ablation} by question type. Table~\ref{tab:full-scaling}
breaks all metrics out across the four \sg{} tiers, exposing the
scaling curve underlying the headline numbers.

\begin{table}[!ht]
\caption{Edge counts and wiring density on GraphRAG-Bench medical.
Edge semantics differ across paradigms (entity co-mention vs.\ typed
causal), so the compactness comparison is primarily about node
vocabulary (Table~\ref{tab:main-results}); Edges/Node illustrates
wiring density, not cross-paradigm comparability.}
\label{tab:graph-sizes}
\centering
\begin{tabular}{lrcc}
\toprule
Framework & Nodes $\downarrow$ & Edges & Edges/Node \\
\midrule
\msgr{}            & 5{,}130 & 10{,}625 & 2.07 \\
\lgr{}             & 4{,}867 & 8{,}104  & 1.67 \\
\fgr{}             & 5{,}807 & 19{,}032 & 3.28 \\
\specialrule{.08em}{.3em}{.3em}
\sg{} (ours) L     & 1{,}583 & 17{,}104 & 10.80 \\
\sg{} (ours) M     & 1{,}064 & 8{,}455  & 7.95 \\
\sg{} (ours) S     & 902     & 8{,}846  & 9.81 \\
\sg{} (ours) XS    & 333     & 978      & 2.94 \\
\bottomrule
\end{tabular}
\end{table}

\begin{table*}[!ht]
\caption{Stratified results on GraphRAG-Bench medical (5 runs).
Correctness for all four question types (FR: Fact Retrieval;
CR: Complex Reasoning; CS: Contextual Summary; CG: Creative
Generation). Coverage for CS and CG; Faithfulness for CG only.
Ablations (\textit{Nodes only}, \textit{In-context}) defined in
Section~\ref{sec:ablation}.}
\label{tab:stratified}
\centering
\footnotesize
\setlength{\tabcolsep}{4pt}
\begin{tabular}{l cccc cc c}
\toprule
 & \multicolumn{4}{c}{Correctness $\uparrow$} & \multicolumn{2}{c}{Coverage $\uparrow$} & Faith.\ $\uparrow$ \\
\cmidrule(lr){2-5}\cmidrule(lr){6-7}\cmidrule(lr){8-8}
Framework & FR & CR & CS & CG & CS & CG & CG \\
\midrule
\nrag{}           & .641 & .555 & .625 & .524 & .613 & .430 & \textbf{.557} \\
\msgr{} local     & .655 & .650 & \textbf{.686} & .602 & .703 & .507 & .198 \\
\msgr{} global    & .551 & .589 & .622 & .583 & \textbf{.716} & .525 & \textemdash \\
\lgr{}            & \textbf{.733} & .633 & .630 & .461 & .540 & .332 & .473 \\
\fgr{}            & .712 & \textbf{.677} & .674 & \textbf{.605} & .589 & .410 & .453 \\
\specialrule{.08em}{.3em}{.3em}
\sg{} L          & .695 & .648 & .656 & .589 & .640 & .530 & .381 \\
\sg{} M          & .703 & .662 & .664 & .595 & .659 & \textbf{.537} & .409 \\
\sg{} S          & .692 & .655 & .645 & .584 & .630 & .516 & .404 \\
\sg{} XS         & .687 & .656 & .640 & .582 & .613 & .491 & .338 \\
\midrule
Nodes only        & .477 & .506 & .436 & .496 & .275 & .317 & .508 \\
In-context        & .539 & .449 & .514 & .432 & .468 & .331 & .216 \\
\bottomrule
\end{tabular}
\end{table*}

\begin{table*}[!ht]
\caption{Scaling analysis on medical: all metrics by question type
and \sg{} tier (5-run means). ROUGE reported for FR/CR; Coverage for
CS/CG; Faithfulness for CG only.}
\label{tab:full-scaling}
\centering
\footnotesize
\setlength{\tabcolsep}{3pt}
\begin{tabular}{l cc cc cc ccc}
\toprule
 & \multicolumn{2}{c}{Fact Retrieval} & \multicolumn{2}{c}{Complex Reas.} & \multicolumn{2}{c}{Ctx.\ Summary} & \multicolumn{3}{c}{Creative Gen.} \\
\cmidrule(lr){2-3}\cmidrule(lr){4-5}\cmidrule(lr){6-7}\cmidrule(lr){8-10}
Tier (nodes) & ROUGE $\uparrow$ & Corr.\ $\uparrow$ & ROUGE $\uparrow$ & Corr.\ $\uparrow$ & Corr.\ $\uparrow$ & Cov.\ $\uparrow$ & Corr.\ $\uparrow$ & Cov.\ $\uparrow$ & Faith.\ $\uparrow$ \\
\midrule
L (1{,}583) & .354 & .695 & .302 & .648 & .656 & .640 & .589 & .530 & .354 \\
M (1{,}064) & \textbf{.360} & \textbf{.703} & .306 & \textbf{.662} & \textbf{.664} & \textbf{.659} & \textbf{.595} & \textbf{.537} & \textbf{.384} \\
S (902)     & .354 & .692 & \textbf{.307} & .655 & .645 & .630 & .584 & .516 & .369 \\
XS (333)    & .350 & .687 & .305 & .656 & .640 & .613 & .582 & .491 & .301 \\
\bottomrule
\end{tabular}
\end{table*}

\FloatBarrier
\section{\epilepsyqa{} detailed results}
\label{app:epilepsy-stratified}

Table~\ref{tab:epilepsy-graph-sizes} lists graph footprints on the
epilepsy corpus; Tables~\ref{tab:epilepsy-stratified}
and~\ref{tab:epilepsy-scaling} report per-question-type breakdowns and
tier scaling on \epilepsyqa{} (375 questions, 5 runs). Unlike
the medical benchmark where quality peaks at the M tier, epilepsy
exhibits a flat scaling curve: L (0.611) leads with XS (0.607), M
(0.604), and S (0.604) all within 0.007 answer correctness. Even the
smallest tier captures sufficient causal structure for effective
retrieval.

\begin{table}[!ht]
\caption{Graph sizes on the epilepsy corpus (same tier labels as
Table~\ref{tab:epilepsy-scaling}). \msgr{} local and global search share
one extracted entity graph. Edge semantics differ across paradigms
(entity co-mention vs.\ typed causal), so the compactness comparison is
primarily about node vocabulary; Edges/Node illustrates wiring density,
not cross-paradigm comparability. Baseline counts are taken from each
framework's built epilepsy index (Microsoft GraphRAG
\texttt{entities}/\texttt{relationships} tables, LightRAG entity and
relation stores, Fast-GraphRAG serialized graph); \sg{} counts are typed
variables and causal edges in the stored graph build.}
\label{tab:epilepsy-graph-sizes}
\centering
\begin{tabular}{lccc}
\toprule
Framework & Nodes $\downarrow$ & Edges & Edges/Node \\
\midrule
\msgr{}      & 4{,}335 & 5{,}642 & 1.30 \\
\lgr{}       & 4{,}148 & 4{,}919 & 1.19 \\
\fgr{}       & 5{,}227 & 9{,}676 & 1.85 \\
\specialrule{.08em}{.3em}{.3em}
\sg{} (ours) L  & 624 & 8{,}708 & 13.96 \\
\sg{} (ours) M  & 468 & 5{,}223 & 11.16 \\
\sg{} (ours) S  & 427 & 6{,}617 & 15.50 \\
\sg{} (ours) XS & 218 & 1{,}311 & 6.01 \\
\bottomrule
\end{tabular}
\end{table}

\begin{table*}[!ht]
\caption{Stratified results on epilepsy (5 runs).
Correctness for all six question types; Coverage for Synthesis only.
Mech: Mechanistic; CF: Counterfactual; MHC: Multi-hop Causal; Syn:
Synthesis; FR: Fact Retrieval; ClinR: Clinical Reasoning.}
\label{tab:epilepsy-stratified}
\centering
\footnotesize
\setlength{\tabcolsep}{4pt}
\begin{tabular}{l cccccc c}
\toprule
 & \multicolumn{6}{c}{Correctness $\uparrow$} & Cov.\ $\uparrow$ \\
\cmidrule(lr){2-7}\cmidrule(lr){8-8}
Framework & Mech & CF & MHC & Syn & FR & ClinR & Syn \\
\midrule
\nrag{}           & .400 & .337 & .364 & .419 & .518 & .449 & .212 \\
\msgr{} local     & \textbf{.623} & \textbf{.589} & .588 & .607 & .618 & .615 & .370 \\
\msgr{} global    & .528 & .581 & .550 & .583 & .610 & .598 & \textbf{.464} \\
\lgr{}            & .338 & .284 & .300 & .366 & .355 & .379 & .144 \\
\fgr{}            & .575 & .559 & .548 & .582 & .631 & .583 & .311 \\
\specialrule{.08em}{.3em}{.3em}
\sg{} L          & .616 & .548 & \textbf{.605} & \textbf{.620} & .722 & \textbf{.630} & .422 \\
\sg{} M          & .605 & .546 & .609 & .613 & .687 & .616 & .425 \\
\sg{} S          & .617 & .544 & .593 & .613 & .713 & .616 & .406 \\
\sg{} XS         & .608 & .557 & .607 & .615 & \textbf{.725} & .619 & .397 \\
\bottomrule
\end{tabular}
\end{table*}

\begin{table}[!ht]
\caption{Scaling analysis on epilepsy (5-run means).}
\label{tab:epilepsy-scaling}
\centering
\begin{tabular}{lccc}
\toprule
Tier (nodes) & Ans.\ Corr.\ $\uparrow$ & ROUGE $\uparrow$ & Coverage $\uparrow$ \\
\midrule
L (624)  & \textbf{.611}{\tiny$\pm$.006} & \textbf{.206}{\tiny$\pm$.002} & .422{\tiny$\pm$.007} \\
M (468)  & .604{\tiny$\pm$.004} & .205{\tiny$\pm$.001} & \textbf{.425}{\tiny$\pm$.009} \\
S (427)  & .604{\tiny$\pm$.006} & .206{\tiny$\pm$.002} & .406{\tiny$\pm$.012} \\
XS (218) & .607{\tiny$\pm$.005} & .201{\tiny$\pm$.001} & .397{\tiny$\pm$.021} \\
\bottomrule
\end{tabular}
\end{table}

\FloatBarrier
\section{Pareto frontier on medical}
\label{app:pareto}

\begin{figure}[!ht]
\centering
\begin{tikzpicture}
\pgfplotsset{
  hcg/.style={mark=*, color=blue!70!black, only marks, mark size=2.6pt},
  msgr/.style={mark=square*, color=orange!85!black, only marks, mark size=2.4pt},
  ent/.style={mark=triangle*, color=red!75!black, only marks, mark size=2.9pt},
  frontier/.style={dashed, color=black!55, line width=0.7pt, mark=none},
  paretoaxis/.style={
    width=0.78\linewidth,
    height=0.36\linewidth,
    grid=both,
    major grid style={line width=.1pt,draw=gray!25},
    minor grid style={line width=.05pt,draw=gray!12},
    tick label style={font=\scriptsize},
    label style={font=\scriptsize},
    title style={font=\scriptsize, yshift=-3pt},
    axis line style={black!60},
    tick style={black!60},
    log ticks with fixed point,
  },
}
\tikzset{
  ptlbl/.style={font=\scriptsize, inner sep=2pt, text=black!75},
}

\begin{axis}[
  xmode=log, log basis x={10},
  xmin=240, xmax=12000,
  ymin=0.555, ymax=0.705,
  xlabel={Graph nodes (log scale)},
  ylabel={Answer correctness},
  title={(a) Correctness vs.\ nodes},
  paretoaxis,
  name=ax1,
  legend to name=paretoLegend,
  legend columns=4,
  legend cell align=left,
  legend style={font=\scriptsize, draw=none, fill=none,
                column sep=14pt, inner sep=1pt},
]
\addplot[frontier] coordinates {(333,0.664) (902,0.668) (1064,0.679) (5807,0.690)};
\addlegendentry{Pareto frontier}
\addplot[hcg] coordinates {(333,0.664) (902,0.668) (1064,0.679) (1583,0.669)};
\addlegendentry{HCG-RAG}
\addplot[ent] coordinates {(4867,0.673) (5807,0.690)};
\addlegendentry{LightRAG / Fast-GraphRAG}
\addplot[msgr] coordinates {(5130,0.654) (5130,0.573)};
\addlegendentry{MS-GraphRAG (local / global)}
\node[ptlbl, anchor=east]  at (axis cs:333,0.664)  {XS\,};
\node[ptlbl, anchor=south] at (axis cs:902,0.668)  {S};
\node[ptlbl, anchor=south] at (axis cs:1064,0.679) {M};
\node[ptlbl, anchor=north] at (axis cs:1583,0.669) {L};
\node[ptlbl, anchor=south] at (axis cs:4867,0.673) {LGR};
\node[ptlbl, anchor=north] at (axis cs:5807,0.690) {FGR};
\node[ptlbl, anchor=west]  at (axis cs:5130,0.654) {\,MSGR-L};
\node[ptlbl, anchor=west]  at (axis cs:5130,0.573) {\,MSGR-G};
\end{axis}

\begin{axis}[
  xmode=log, log basis x={10},
  xmin=240, xmax=12000,
  ymin=0.450, ymax=0.665,
  xlabel={Graph nodes (log scale)},
  ylabel={Coverage},
  title={(b) Coverage vs.\ nodes},
  paretoaxis,
  at={(ax1.outer south west)},
  yshift=-12pt,
  anchor=outer north west,
  name=ax2,
]
\addplot[frontier] coordinates {(333,0.570) (902,0.589) (1064,0.616) (5130,0.632) (5130,0.649)};
\addplot[hcg] coordinates {(333,0.570) (902,0.589) (1064,0.616) (1583,0.601)};
\addplot[ent] coordinates {(4867,0.466) (5807,0.525)};
\addplot[msgr] coordinates {(5130,0.632) (5130,0.649)};
\node[ptlbl, anchor=east]  at (axis cs:333,0.570)  {XS\,};
\node[ptlbl, anchor=south] at (axis cs:902,0.589)  {S};
\node[ptlbl, anchor=south] at (axis cs:1064,0.616) {M};
\node[ptlbl, anchor=north] at (axis cs:1583,0.601) {L};
\node[ptlbl, anchor=west]  at (axis cs:4867,0.466) {\,LGR};
\node[ptlbl, anchor=west]  at (axis cs:5807,0.525) {\,FGR};
\node[ptlbl, anchor=west]  at (axis cs:5130,0.632) {\,MSGR-L};
\node[ptlbl, anchor=west]  at (axis cs:5130,0.649) {\,MSGR-G};
\end{axis}

\node[anchor=north, yshift=-4pt]
  at ($(ax2.outer south west)!0.5!(ax2.outer south east)$)
  {\pgfplotslegendfromname{paretoLegend}};

\end{tikzpicture}
\caption{Quality--size frontier on GraphRAG-Bench medical (5-run
means, GLM-4.7 judge). Dashed lines trace the non-dominated set in
each 2-D projection of the (correctness, coverage, nodes) space,
using node count as a graph-footprint proxy. \emph{Top}: \sg{} XS,
S, and M form the low-node arc of the correctness--nodes frontier;
\fgr{} anchors the high-correctness end (0.690). \lgr{} is strictly
dominated by \sg{}~M (lower correctness, lower coverage, more
nodes); \sg{}~L is dominated by \sg{}~M on both axes, signaling
diminishing returns past M. \emph{Bottom}: \sg{} XS, S, M form the
low-node portion of the coverage--nodes frontier, transitioning to
\msgr{}'s higher-coverage, higher-node regime; \lgr{}, \fgr{}, and
\sg{}~L are dominated on this projection.}
\label{fig:pareto}
\end{figure}
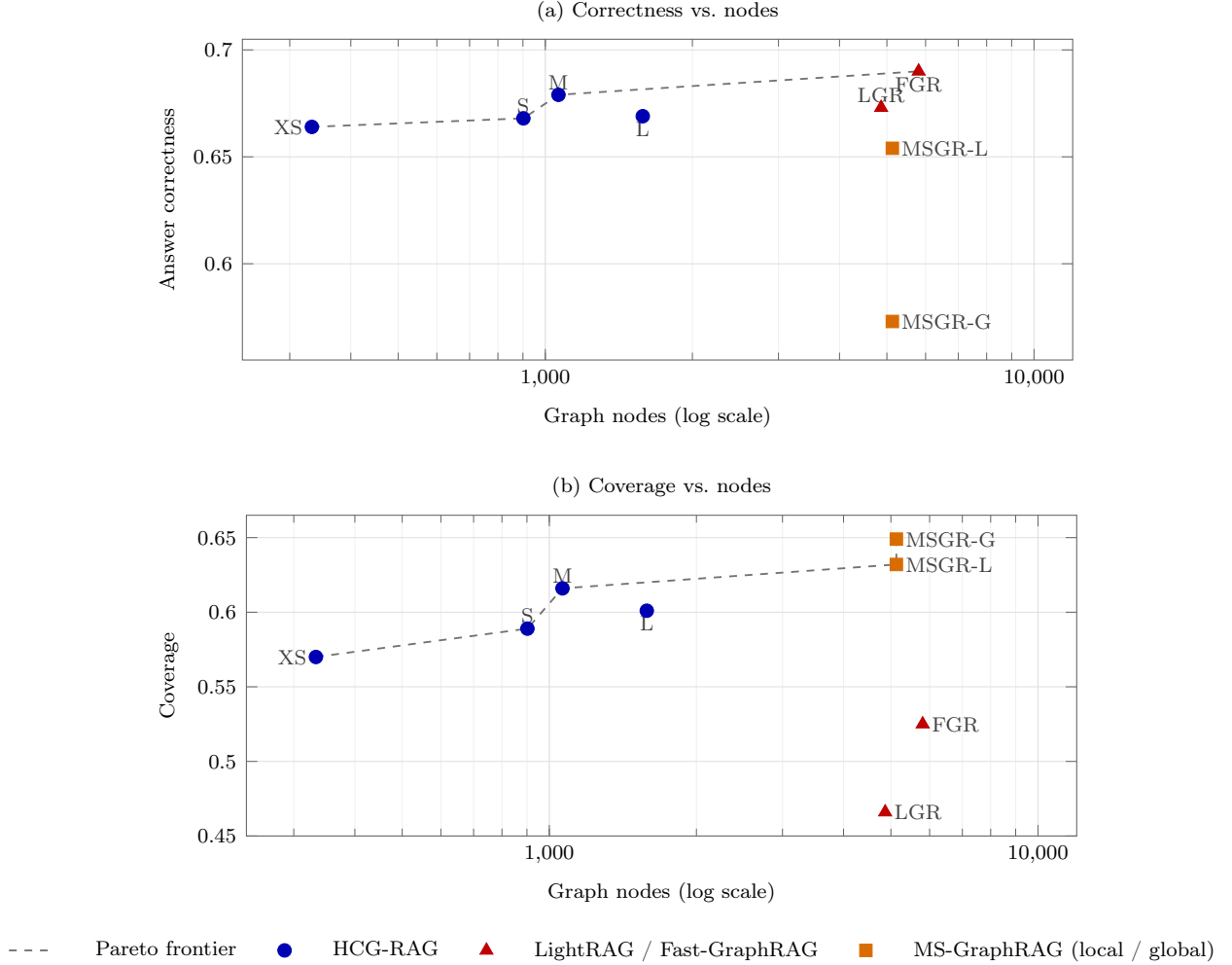

Figure~\ref{fig:pareto} visualizes the quality--size trade-off on
GraphRAG-Bench medical, using node count as a graph-footprint proxy and
projecting the (correctness, coverage, nodes) space onto two 2-D planes
(correctness vs.\ nodes and coverage vs.\ nodes). \fgr{} sits on the
frontier as the high-correctness extreme (0.690 correctness, 0.525
coverage, 5{,}807 nodes); \sg{}~M occupies the high-coverage, low-node
trade-off (0.679 correctness, 0.616 coverage, 1{,}064 nodes), trading
1.1\,pp correctness for $+9.1$\,pp coverage and a $5.5\times$ smaller
graph. \lgr{} is strictly dominated by \sg{}~M on all three axes
($+0.6$\,pp correctness, $+15.0$\,pp coverage, $4.6\times$ fewer nodes).
The three smaller \sg{} tiers (XS, S, M) sit on the frontier in both
projections, occupying the low-node arc that no other system reaches;
\msgr{} occupies a complementary high-coverage, high-node regime
(0.649 coverage, 5{,}130 nodes for global). The L tier (1{,}583 nodes)
is dominated by M on correctness ($-$1.0\,pp) and coverage
($-$1.5\,pp), signalling diminishing returns past 1k nodes.

\section{Paired bootstrap significance}
\label{app:significance}

Run-to-run standard deviation in Tables~\ref{tab:main-results}
and~\ref{tab:epilepsy-results} reflects API/decoding noise across
the five inference runs at temperature~0; it is not a test of
significance across the benchmark. To address that gap, we compute
paired bootstrap confidence intervals on per-question answer
correctness, using the per-question scores already produced by the
GLM-4.7 judge for every system on every benchmark.

\paragraph{Procedure.}
For each pair $(A, B)$ on a benchmark, we average each system's
per-question correctness over the five runs to obtain a paired
score vector, restrict to the intersection of question IDs, and
take a paired bootstrap on the per-question difference
$\Delta_q = s^A_q - s^B_q$ ($B = 10{,}000$ resamples, seed $=42$).
We report mean $\Delta$, the 2.5/97.5 percentile interval, and a
two-sided $p$-value (twice the smaller bootstrap-mean tail mass).

\paragraph{Headline comparisons.}
Table~\ref{tab:bootstrap} reports significance for the comparisons
referenced in the main text. The pattern reinforces a more
calibrated reading: \sg{}'s wins over flat entity-relation methods
(\fgr{} on epilepsy, \lgr{} on epilepsy and medical, \nrag{} on
medical) are statistically significant; the closest pairs in the
top cluster (\sg{} tiers vs.\ \msgr{}~local on epilepsy) are
statistical ties; and \sg{}'s 1.1\,pp medical gap behind \fgr{} is
significant rather than within noise, supporting the framing of
the medical result as a $5\times$-graph-size trade rather than
parity.

\begin{table}[!ht]
\caption{Paired bootstrap on per-question answer correctness
(B${=}10{,}000$ resamples, seed${=}42$). $\Delta$ is the mean
per-question difference $A - B$; CI is the 95\% percentile
interval; $p$ is two-sided. Per-pair $n$ is the size of the
question-ID intersection across the five runs; the medical \sg{}~M
vs.\ \lgr{} pair drops to 2061 because one question has no \lgr{}
output across all five runs and is excluded from that pair only.
Non-significant ties (\sg{}~tiers vs.\ \msgr{}~local on epilepsy)
indicate failure to reject the null at $\alpha{=}.05$, not formal
equivalence; tightness of the 95\% CI (e.g.\ \sg{}~L vs.\ \msgr{}
local: $[-.006, +.016]$) bounds the practical magnitude of any
remaining effect.}
\label{tab:bootstrap}
\centering
\footnotesize
\begin{tabular}{l l r l r r c}
\toprule
Benchmark & A vs.\ B & $\Delta$ & 95\% CI & $p$ & $n$ & Sig.\ @ .05 \\
\midrule
medical   & \sg{}~M vs.\ \nrag{}        & $+.070$ & $[+.059, +.082]$ & $<.001$ & 2062 & yes \\
medical   & \sg{}~M vs.\ \msgr{}~local  & $+.025$ & $[+.016, +.034]$ & $<.001$ & 2062 & yes \\
medical   & \sg{}~M vs.\ \lgr{}         & $+.006$ & $[-.005, +.018]$ & $.27$   & 2061 & no  \\
medical   & \sg{}~M vs.\ \fgr{}         & $-.011$ & $[-.020, -.002]$ & $.02$   & 2062 & yes \\
medical   & \sg{}~L vs.\ \fgr{}         & $-.020$ & $[-.030, -.011]$ & $<.001$ & 2062 & yes \\
medical   & \sg{}~XS vs.\ \fgr{}        & $-.025$ & $[-.035, -.015]$ & $<.001$ & 2062 & yes \\
\midrule
epilepsy  & \sg{}~L vs.\ \lgr{}         & $+.277$ & $[+.256, +.299]$ & $<.001$ & 375  & yes \\
epilepsy  & \sg{}~L vs.\ \fgr{}         & $+.041$ & $[+.030, +.053]$ & $<.001$ & 375  & yes \\
epilepsy  & \sg{}~L vs.\ \msgr{}~local  & $+.005$ & $[-.006, +.016]$ & $.41$   & 375  & no  \\
epilepsy  & \sg{}~XS vs.\ \msgr{}~local & $+.001$ & $[-.010, +.013]$ & $.82$   & 375  & no  \\
epilepsy  & \sg{}~M vs.\ \msgr{}~local  & $-.002$ & $[-.013, +.010]$ & $.77$   & 375  & no  \\
epilepsy  & \sg{}~L vs.\ \sg{}~XS       & $+.003$ & $[-.004, +.011]$ & $.38$   & 375  & no  \\
\midrule
novel     & \sg{}~S vs.\ \fgr{}         & $-.072$ & $[-.083, -.061]$ & $<.001$ & 2009 & yes \\
novel     & \sg{}~S vs.\ \lgr{}         & $-.055$ & $[-.067, -.043]$ & $<.001$ & 2009 & yes \\
novel     & \sg{}~S vs.\ \nrag{}        & $-.019$ & $[-.031, -.008]$ & $.001$  & 2009 & yes \\
\bottomrule
\end{tabular}
\end{table}

\paragraph{Cross-stratum stability on epilepsy (hop-heavy vs.\
lookup-heavy).}
To test whether \sg{}'s advantage over flat entity-relation
retrieval concentrates on multi-hop questions, we group epilepsy
questions into a hop-heavy slice (Multi-hop Causal,
Counterfactual, Synthesis; $n{=}210$) and a lookup-heavy slice
(Fact Retrieval, Mechanistic; $n{=}108$) and run paired bootstrap
on each. \sg{}~L significantly outperforms \fgr{} on both slices
($\Delta{=}{+}.038$ and ${+}.044$; both $p{<}.001$) and \lgr{} on
both ($\Delta{>}{+}.28$; $p{<}.001$). Versus \msgr{}~local it is
tied on both ($p{=}.51$ and $p{=}.94$). The advantage of structured
retrieval over flat entity-relation methods holds whether the
question requires chained mechanism reasoning or a typed-vocabulary
lookup (it is not concentrated on a single question type).

\FloatBarrier
\section{Cross-family judge robustness (Kimi-K2.5)}
\label{app:cross-judge}

To validate that the GLM-4.7 judge column is not a single-judge
artefact, we re-evaluate every paper-era inference prediction under a
second judge from a different vendor family: \textbf{Kimi-K2.5}
(Moonshot AI, on AWS Bedrock).

\paragraph{Per-system rank correlation (GLM \(\leftrightarrow\) Kimi).}

\begin{table}[!ht]
\caption{Spearman $\rho$ and Kendall $\tau$ between GLM-4.7 and
Kimi-K2.5 on per-system mean answer correctness, per benchmark.
Paired bootstrap, B${=}10{,}000$, seed${=}42$.}
\label{tab:cross-judge-rank}
\centering
\footnotesize
\begin{tabular}{lccr}
\toprule
Benchmark & Spearman $\rho$ (95\% CI) & Kendall $\tau$ (95\% CI) & $n_{\text{systems}}$ \\
\midrule
Medical  & $0.93$ $[0.73, 0.98]$ & $0.83$ $[0.61, 0.94]$ & 9 \\
Epilepsy & $0.71$ $[0.45, 0.93]$ & $0.60$ $[0.38, 0.82]$ & 10 \\
Novel    & $1.00$ $[1.00, 1.00]$ & $1.00$ $[1.00, 1.00]$ & 4 \\
\bottomrule
\end{tabular}
\end{table}

\paragraph{Side-by-side per-system means.}
Table~\ref{tab:cross-judge-side} reports per-system means under both
judges on each benchmark. Kimi is on average stricter than GLM
($\bar{\Delta}{=}-0.012$ medical, $-0.014$ novel), with epilepsy as
an exception where Kimi rates several systems more leniently
($\bar{\Delta}{=}+0.014$, with \msgr{} variants gaining the most).

\begin{table}[!ht]
\caption{Per-system answer correctness under GLM-4.7 vs.\ Kimi-K2.5.
$\Delta = $ Kimi $-$ GLM; positive $=$ Kimi more lenient.}
\label{tab:cross-judge-side}
\centering
\footnotesize
\begin{tabular}{lcccc cccc cccc}
\toprule
& \multicolumn{4}{c}{\textbf{Medical}} & \multicolumn{4}{c}{\textbf{Epilepsy}} & \multicolumn{4}{c}{\textbf{Novel}} \\
\cmidrule(lr){2-5}\cmidrule(lr){6-9}\cmidrule(lr){10-13}
System & GLM & Kimi & $\Delta$ && GLM & Kimi & $\Delta$ && GLM & Kimi & $\Delta$ & \\
\midrule
\nrag{}      & .608 & .593 & $-.015$ && .397 & .388 & $-.009$ && .544 & .554 & $+.011$ & \\
\msgr{} local  & .654 & .631 & $-.023$ && .606 & .634 & $+.029$ && --   & --   & --     & \\
\msgr{} global & .573 & .549 & $-.024$ && .562 & .609 & $+.048$ && --   & --   & --     & \\
\lgr{}       & .673 & .655 & $-.018$ && .334 & .322 & $-.012$ && .579 & .581 & $+.002$ & \\
\fgr{}       & .690 & .662 & $-.027$ && .569 & .566 & $-.003$ && .596 & .603 & $+.007$ & \\
\sg{}~L      & .669 & .660 & $-.009$ && .611 & .611 & $+.001$ && --   & --   & --     & \\
\sg{}~M      & .679 & .666 & $-.013$ && .604 & .605 & $+.001$ && --   & --   & --     & \\
\sg{}~S      & .668 & .657 & $-.011$ && .604 & .605 & $+.001$ && .524 & .532 & $+.008$ & \\
\sg{}~XS     & .664 & .653 & $-.012$ && .607 & .600 & $-.007$ && --   & --   & --     & \\
\bottomrule
\end{tabular}
\end{table}

\paragraph{Significance-flag survival.}
Replicating the paired bootstrap of Table~\ref{tab:bootstrap} under
Kimi (same per-question procedure, BH-FDR-adjusted $q$-values within
family) preserves the headline verdict on five of seven pairs: on
medical, \sg{}~M vs.\ \lgr{} stays a tie and vs.\ \msgr{}~local and
\nrag{} stay wins ($q{<}.001$); on epilepsy, \sg{}~L vs.\ \fgr{} and
vs.\ \lgr{} stay wins ($q{<}.001$ under both judges). The two shifts
pull in opposite directions: medical \sg{}~M vs.\ \fgr{} softens from
lose ($q{=}.025$ GLM) to tie ($q{=}.44$ Kimi), while epilepsy \sg{}~L
vs.\ \msgr{}~local moves from tie ($q{=}.42$) to a Kimi loss
($q{<}.001$), leaving the aggregate narrative roughly neutral.

\paragraph{What survives, what shifts.}
The categorical claim that organized methods (\sg{} and \msgr{})
significantly outperform flat entity-relation retrieval on epilepsy
is robust under both judges (\sg{}~L vs.\ \fgr{} $\Delta{=}+.041$ GLM,
$+.045$ Kimi, both $q{<}.001$). On medical, \sg{}~M's competitive
position with the strongest entity-relation system is
\emph{strengthened} under Kimi (M vs.\ \fgr{} moves from $q{=}.025$
to $q{=}.44$). The single result that does not survive Kimi as
worded is the epilepsy tie between \sg{}~L and \msgr{}~local: under
Kimi, \msgr{}~local leads by $2.3$\,pp ($q{<}.001$). Notably,
\sg{}'s epilepsy scores are judge-stable across all four tiers
($|\Delta| {\leq} .007$), while \msgr{}~local's score swings $+.029$,
indicating the cross-judge disagreement reflects judge-dependent
generosity toward \msgr{}'s community-summary outputs rather than
instability in \sg{}.

\section{Qualitative case studies}
\label{app:qualitative}

We examine four medical Complex-Reasoning questions where the causal
graph produces qualitatively different answers from entity-relation
baselines. These are representative of the 32 questions where \sg{}
exceeds \fgr{} by more than 0.3 answer correctness; all involve
multi-hop causal chains that flat entity retrieval cannot reconstruct.

\paragraph{Hodgkin lymphoma diagnosis (visualized in Figure~\ref{fig:retrieval}).}
\emph{Q: A 25-year-old presents with swollen lymph nodes in the upper
body. What cancer is most likely?} Ground truth: Hodgkin lymphoma
(common ages 15--30, starts in upper body nodes). \sg{} answers
``Hodgkin lymphoma, because in a 25-year-old, swollen upper-body lymph
nodes are a classic presentation'' (0.969). \fgr{} answers ``breast
cancer'' (0.380), retrieving passages about breast masses rather than
following the age-presentation-diagnosis chain. The causal graph
connects age-stratified risk factors to lymphoma subtypes via directed
edges, enabling the correct differential.

\paragraph{Bladder cancer treatment.}
\emph{Q: Which treatments are used for non-muscle-invasive bladder
cancer?} Ground truth: TURBT and intravesical therapy (BCG or
chemotherapy). \sg{} answers ``treated with transurethral resection of
the bladder tumor (TURBT), often followed by intravesical therapy such
as chemotherapy or BCG'' (0.982). \fgr{} answers ``commonly treated
with surgery, chemotherapy, radiation therapy, immunotherapy'' (0.366),
a generic oncology response that misses the stage-specific protocol.
The causal graph links cancer staging variables to treatment response
variables, retrieving evidence about the specific TURBT pathway.

\paragraph{Biomarker interpretation.}
\emph{Q: What is the role of p16 protein in oropharyngeal squamous cell
carcinoma?} Ground truth: p16 overexpression identifies HPV-mediated
SCC. \sg{} answers ``p16 is a surrogate biomarker used to identify
HPV-positive oropharyngeal squamous cell carcinoma'' (0.983). \fgr{}
describes p16 vaguely as a ``biomarker test'' without connecting it to
the HPV mechanism (0.394). The causal graph encodes the
biomarker$\to$subtype$\to$prognosis chain, so BFS from a p16-matched
node retrieves the HPV connection.

\paragraph{Treatment decision reasoning.}
\emph{Q: Why don't all follicular lymphoma patients need immediate
treatment?} \sg{} answers ``because many cases are indolent and
low-risk, so they can be safely monitored with watchful waiting''
(0.879). \fgr{} produces a generic statement about treatment decisions
(0.346). The causal graph links disease aggressiveness to treatment
urgency, retrieving evidence about the indolent subtype.

\medskip\noindent
Across these cases, \sg{}'s advantage stems from the same mechanism:
BFS traversal follows causal pathways (risk factor$\to$diagnosis,
staging$\to$treatment, biomarker$\to$subtype) and collects evidence
along the chain. Entity-relation retrieval surfaces topically related
passages but cannot reconstruct the multi-hop reasoning path.
Qualitative case studies from \epilepsyqa{} are deferred to
future work pending expert clinical review of retrieved causal chains.

\section{Experimental details}
\label{app:experimental}

\paragraph{Model configuration.}
All frameworks use \texttt{gpt-5.4-mini} for generation and
\texttt{text-embedding-3-small} for embeddings, accessed via the OpenAI
API. Temperature is set to 0 for reproducibility.

\paragraph{Competitor configuration.}
\fgr{} and \lgr{} are run in isolated Python virtual environments to
avoid dependency conflicts (they pin different versions of the
\texttt{openai} package). \msgr{} is run via direct import. All
competitors use their default configurations except for the model and
embedding model, which are standardized.

\paragraph{Evaluation.}
The LLM-as-judge evaluator uses GLM-4.7~\citep{glm47} via a LiteLLM
proxy~\citep{litellm} fronting AWS Bedrock; metrics are from
GraphRAG-Bench~\citep{graphragbench}: answer correctness, coverage
score, faithfulness, and ROUGE score. Metrics are computed per-question
and aggregated per question type. We use a non-OpenAI judge to avoid
same-family bias when evaluating \texttt{gpt-5.4-mini}--extracted answers;
the embedding component of answer correctness remains
\texttt{text-embedding-3-small} for direct comparability with prior
GraphRAG-Bench results.

\paragraph{Hardware.}
All experiments were executed on an AWS \texttt{c6a.4xlarge} instance
(16~vCPUs, AMD EPYC 7R13; 32\,GB RAM; 200\,GB EBS; Ubuntu 24.04) in
\texttt{us-east-1}. No GPU was used: graph construction and answer
generation are API-bound (OpenAI), so local computation is minimal.

\paragraph{Domain specification provenance.}
All domain specifications used in the benchmark (outcome
configuration and generated schemas) were produced end-to-end by the
automated pipeline (\S\ref{sec:domspec}) with no human review or
editing. \sg{}'s domain-specification stage thus produces usable
causal schemas without expert intervention. The
pipeline is also designed to be human-editable: a domain expert can
revise or author the outcome configuration to encode prior causal
knowledge, potentially improving graph quality beyond what fully
automated generation achieves.

\section{Graph construction and query cost}
\label{app:cost}

Table~\ref{tab:build-cost} summarizes one-time build cost and recurring
\textbf{GraphRAG-Bench} query cost on medical
(\(2{,}062\) questions).  For \sg{} API columns we report \textbf{one}
embedding per answered question plus
\textbf{one} extractor LLM.

\begin{table*}[!ht]
\caption{Build and query cost on the medical benchmark (2{,}062
questions). Token counts are per query. \lgr{} caches LLM responses
after the first run, reducing subsequent query latency to $<$1\,s;
uncached numbers are shown. \msgr{} local and global share a single
build phase; global search map-reduces over all community reports,
yielding $\sim$196 LLM calls per query.}
\label{tab:build-cost}
\centering
\footnotesize
\setlength{\tabcolsep}{4pt}
\begin{tabular}{l rrr rr cc rr}
\toprule
 & \multicolumn{3}{c}{Build (one-time)} & \multicolumn{2}{c}{Query} & \multicolumn{2}{c}{API calls/q} & \multicolumn{2}{c}{Tokens/q} \\
\cmidrule(lr){2-4}\cmidrule(lr){5-6}\cmidrule(lr){7-8}\cmidrule(lr){9-10}
Framework & Time $\downarrow$ & LLM $\downarrow$ & Emb. $\downarrow$ & Batch $\downarrow$ & Latency $\downarrow$ & Emb $\downarrow$ & LLM $\downarrow$ & Emb $\downarrow$ & LLM $\downarrow$ \\
 & (min) & calls & calls & (s/q) & (s/q) & calls/q & calls/q & tok/q & prompt tok/q \\
\midrule
\nrag{}     & $<$1 & 0     & 3      & 0.3 & 1.6 & 1 & 1 & 10 & 7{,}500 \\
\lgr{}      & 34   & 4{,}423 & 15{,}889 & 0.3 & 2.5 & 1 & 2 & 45 & 26{,}545 \\
\fgr{}      & 13   & 2{,}882 & 182    & 0.8 & 2.7 & 1 & 2 & 27 & 9{,}054 \\
\midrule
\msgr{} local  & \multirow{2}{*}{28} & \multirow{2}{*}{17{,}638} & \multirow{2}{*}{33} & 1.2 & 3.8 & 2 & 2 & 21 & 18{,}664 \\
\msgr{} global &                     &                           &                      & 9.5 & 42.8 & 0 & 196 & 0 & 2{,}339{,}848 \\
\specialrule{.08em}{.3em}{.3em}
\sg{} (ours) L  & 42   & 2{,}214 & 1{,}344 & 10.8 & 12.6 & 1 & 1 & 11 & 10{,}730 \\
\sg{} (ours) M  & 27   & 1{,}167 & 387  & 3.4 & 5.8 & 1 & 1 & 11 & 12{,}080 \\
\sg{} (ours) S  & 26   & 833   & 90     & 4.0 & 6.3 & 1 & 1 & 11 & 11{,}430 \\
\sg{} (ours) XS & 8    & 131   & 53     & 0.2 & 3.2 & 1 & 1 & 11 & 12{,}700 \\
\bottomrule
\end{tabular}
\end{table*}

\msgr{}'s build is the most LLM-intensive: 17{,}638~calls in 28~min,
dominated by 16{,}354~entity/relationship description summarizations
and 1{,}085~community reports across five hierarchy levels. This is
$4\times$ more LLM calls than \lgr{} (4{,}423) and $6\times$ more
than \fgr{} (2{,}882). \sg{}'s build cost is comparable to the
lighter competitors: 26~minutes and 833~LLM calls for the S~tier;
the XS tier builds in 8~minutes with only 131~LLM calls, strictly
cheaper than \fgr{} (13~min, 2{,}882~calls). The L~tier
(42~min) trades longer build time for richer graph structure but is
dominated by M on the quality--size Pareto frontier
(Section~\ref{sec:scaling}); the M/S/XS tiers are the recommended
operating points. Beyond raw counts, the qualitative difference
matters: ${\sim}$1k typed, human-readable nodes versus 5.8k opaque
entities enables expert audit and correction, and on epilepsy
\sg{} outperforms \fgr{} by 3.5--4.2\,pp despite far fewer nodes
(Table~\ref{tab:epilepsy-results}).
Per-query API cost is \emph{lower} for \sg{}: a single LLM call
(answer generation from retrieved evidence), while \lgr{} and \fgr{}
each make 2, and \msgr{} local makes 2. \msgr{} global is an outlier
at $\sim$196~LLM calls per query (map-reduce over all community
reports), yielding the highest per-query latency (42.8\,s) but
also the highest coverage (Table~\ref{tab:main-results}).

\paragraph{Per-query token cost.}
Among entity-relation systems, \fgr{} is the most token-efficient
(9.0k prompt tok/q) while \lgr{} sends the most (26.5k) due to a
large entity-store context window. \sg{} falls between them
(11--13k for S/M tiers), with the L~tier slightly lower because BFS
retrieves fewer evidence chunks per query.  \msgr{}'s two methods sit
at the extremes: \msgr{} local at 18.7k tok/q is comparable to
\sg{}~M and \lgr{}; \msgr{} global is the outlier at
$\approx 2.34$M~tok/q, roughly $\mathbf{200\times}$ more prompt
tokens per query than \sg{}~M and $260\times$ more than \fgr{}.  The
inflation comes from map-reducing over $\sim$196~community reports
($\approx$12k~tokens each) for every query, which also produces the
42.8\,s/q latency in the same row.

The key query-time cost difference for \sg{} is \emph{latency} rather
than API spend.  A single \sg{} query takes 3.2--12.6\,s depending on
graph size, with only $\approx$1.5--2.0\,s spent on the API call;
the rest is graph traversal and cone merging (CPU-bound). The
XS~tier (3.2\,s) is comparable to \fgr{} (2.7\,s) at similar quality
(Table~\ref{tab:main-results}), while larger tiers trade latency for
richer evidence retrieval.

\end{document}